\documentclass[5p,times,twocolumn]{elsarticle}
\usepackage{amssymb}
\usepackage{amsmath}
\usepackage{multirow} 
\usepackage{bm}
\usepackage{times}
\usepackage{soul}
\usepackage{url}
\usepackage[utf8]{inputenc}
\usepackage[small]{caption}
\usepackage{subcaption}
\usepackage{graphicx}
\usepackage{amsthm}
\usepackage{booktabs}
\usepackage{algorithm}
\usepackage{algorithmic}
\usepackage{xcolor} 
\usepackage{makecell}
\usepackage[switch]{lineno}
\usepackage{pifont}
\newcommand{\cmark}{\ding{51}}%
\newcommand{\xmark}{\ding{55}}%
\newcommand{\pmark}{$\circ$}%
\usepackage{braket}
\usepackage{mathtools}

\usepackage[hidelinks]{hyperref}
\journal{}
\begin{document}

\begin{frontmatter}

\title{Self-Evolving AI for Humanoids: Mechanisms, Safety, and Evaluation of Post-Deployment Self-Improvement}

\author[khu]{Loc X. Nguyen}
\ead{xuanloc088@khu.ac.kr}          

\author[khu]{Avi Deb Raha}
\ead{avi@khu.ac.kr}   

\author[khua]{Huy Q. Le}
\ead{quanghuy69@khu.ac.kr}   

\author[khu]{Eui-Nam Huh}
\ead{johnhuh@khu.ac.kr}

\author[ntu]{Dusit Niyato}
\ead{dniyato@ntu.edu.sg}

\author[khu]{Choong Seon Hong\corref{cor1}}
\ead{cshong@khu.ac.kr}            
\cortext[cor1]{Corresponding author.}

\affiliation[khu]{organization={School of Computing, Kyung Hee University}, city={Yongin-si}, postcode={17104}, country={Republic of Korea}}
\affiliation[khua]{organization={G-LAMP NEXUS Institute, Kyung Hee University}, city={Yongin-si}, postcode={17104}, country={Republic of Korea}}

\affiliation[ntu]{organization={College of Computing and Data Science, Nanyang Technological University}, country={Singapore}}

\begin{abstract}

Humanoid robots are becoming an important part of embodied artificial intelligence, driven by advances in reinforcement learning for locomotion, world models for prediction, and vision-language-action models for general control. However, most of these systems remain static after deployment. A policy is trained offline for a fixed objective and then frozen, even though the tasks, environments, and robot bodies keep drifting over time. An emerging paradigm of self-evolving agents aims to address this problem by allowing systems to improve from their own post-deployment experience. Since most existing studies focus on disembodied software agents, this survey examines how self-evolution changes when an agent has a physical body. We first define self-evolution for humanoids and represent a deployed robot using a state tuple that includes its policy, perception, memory, workflow, and body. This state is updated by an evolution operator in a slow outer loop with a lifelong objective. We then organize the literature into four complementary mechanisms of self-evolution, presented in increasing order of autonomy: self-learning, self-adaptation, self-optimization, and self-generation. Since changes to a humanoid can introduce physical hazards, we treat safety and uncertainty as key design dimensions of the evolution operator, and further formulate admissible evolution as a constraint enforced by a world-model verification gate within a human-oversight envelope. Finally, we present that evaluation should track the robot’s evolving trajectory rather than a fixed checkpoint, and we identify the lack of a benchmark designed specifically for self-evolving humanoids. Moreover, we outline open challenges spanning AI algorithms, on-board systems, and governance.
\end{abstract}


\end{frontmatter}

\section{Introduction}
\label{sec:intro}

\begin{figure*}[t]
\centering
\includegraphics[width=.95\textwidth]{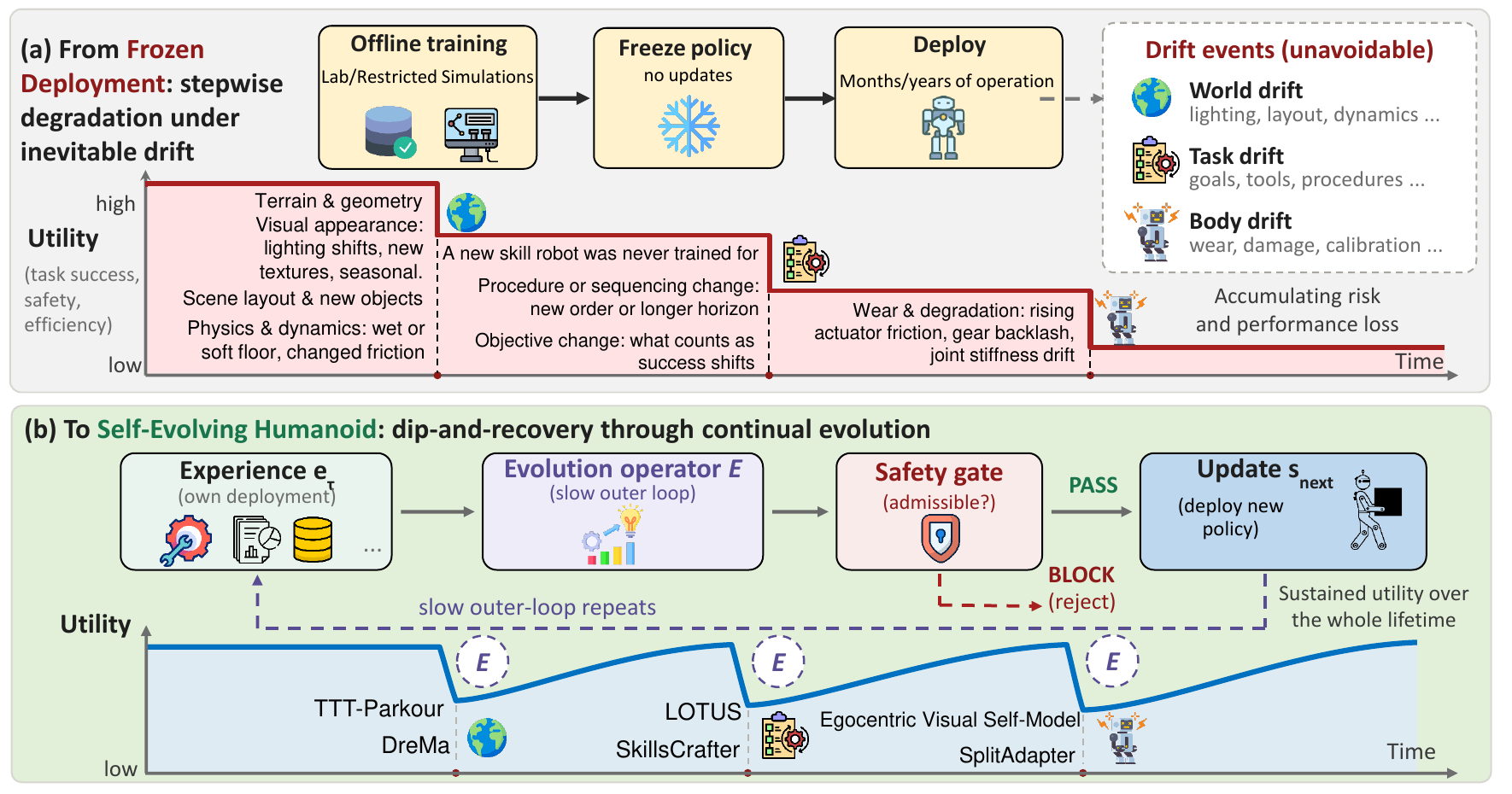}
\caption{From frozen deployment to self-evolution. (a) The conventional pipeline trains a policy offline, freezes it, and deploys it for a long period (months/years) while the world, tasks, and robot body change. As these changes occur, the system's utility can gradually degrade. (b) A self-evolving humanoid uses its deployment experience to update itself through an evolution process. Examples include TTT-Parkour~\citep{tttparkour2026}, DreMa~\citep{dreamtomanipulate2025}, LOTUS~\citep{wan2024lotus}, SkillsCrafter~\citep{skillscrafter2026}, Egocentric model~\citep{hu2025selfmodel}, and SplitAdapter~\citep{splitadapter2026}.}
\label{fig:motivation}

\end{figure*}

Humanoid robots have attracted growing attention as advances in perception, reasoning, and dexterous interaction have made more capable physical AI systems possible. This progress has been driven by three model families: reinforcement learning (RL) for locomotion and whole-body control~\citep{berkeleyhumanoid2024,exbody2_2024,fu2024humanplus}, world models for prediction and planning~\citep{hafner2025dreamerv3,zhang2025dreamvla}, and vision-language-action (VLA) models for general language-conditioned control~\citep{brohan2023rt2,kim2024openvla,black2024pi0}. Although these model families have advanced rapidly, they share a common limitation: they are typically trained offline to optimize a predefined objective and then remain frozen after deployment. This paradigm works well in controlled laboratory settings, but becomes  limiting in the real world, where tasks, environments, and even the robot’s physical condition can change over years of operation. To address this gap, self-evolving agents have recently emerged as a promising solution by allowing the systems to improve from post-deployment experience~\citep{gao2026selfevolving,zhang2026dgm}. However, these studies focus on disembodied software agents. This gap motivates our survey of self-evolving agents with physical bodies, as shown in Figure~\ref{fig:motivation}.

\subsection{Motivation for Self-Evolution}
\label{sec:motivation}

The work of \citep{cao2024humanoid} describes the evolution of humanoid robots in three stages: human-looking, human-like, and human-level. Most existing systems are still at the human-like stage, with strong capabilities in perception, locomotion, and manipulation. Reaching the human-level stage, however, requires more than improving task performance. Specifically, humanoids must also continue to acquire new knowledge and skills after deployment as their operating conditions change. This requirement exposes a fundamental limitation of current humanoid systems, whose learned policies and foundation models are typically trained offline and remain frozen during deployment~\citep{ding2025humanoidvla}. Although these static models can perform well in controlled settings, their performance can degrade when the distribution shifts because of new environments, new tasks, new users, or changes in the robot's own body. Humanoids must therefore adapt continually to operate in dynamic physical environments. Accordingly, research is shifting from static foundation models toward lifelong, self-evolving intelligence that improves through ongoing interaction with the real world~\citep{fang2025selfevolving,wang2026self}.

Humanoid robots are expected to operate for years in environments that continue to change after deployment. As summarized in Table~\ref{tab:drivers}, these changes come from four sources: the changing world, the changing robot, changing knowledge, and changing intelligence requirements. Together, these drivers create persistent shifts that can reduce the effectiveness of policies trained offline. Addressing this gap requires humanoid systems that can continually adapt, acquire new capabilities, and improve existing behaviors throughout their operational lifetime~\citep{meng2025legion}. Every second of operation, a humanoid generates images, video, force and tactile measurements, proprioception, successful and failed trajectories, and human feedback~\citep{Self-organized}. This provides a rich source of experience for post-deployment improvement. However, most existing robots discard much of this experience~\citep{wan2024lotus}, while a self-evolving humanoid can use it to continually improve its models. The challenge is that a single humanoid relies on many learned models for locomotion, control, internal simulation and prediction, semantic perception, planning, and reasoning. Updating all of these models manually is therefore impractical.

\begin{table*}[t]
\centering
\caption{Four drivers that keep changing after deployment and create the distribution shift.}
\label{tab:drivers}
\scalebox{0.85}{
\begin{tabular}{p{0.18\textwidth} p{0.9\textwidth}}
\toprule
\textbf{Driver} & \textbf{Why self-evolution is needed} \\
\midrule
Changing World &
Dynamic environments, new objects, evolving tasks, and shifting user needs create continual distribution shift. \\

Changing Robot &
Hardware degradation, sensor drift, wear, maintenance, and morphology changes require ongoing adaptation. \\

Changing Knowledge &
Foundation models, semantic understanding, and planning capabilities must remain current as the world evolves. \\

Changing Intelligence &
Long-term autonomy demands continual skill acquisition, reflection on experience, self-correction, and open-ended improvement rather than fixed capabilities. \\
\bottomrule
\end{tabular}}
\end{table*}

\begin{table*}[t]
\centering
\caption{Positioning against related surveys. \cmark~= treated as a central concern; \pmark~= partial or incidental; \xmark~= not addressed.}
\label{tab:surveys}
\small
\scalebox{0.85}{
\begin{tabular}{lccccc}
\toprule
\textbf{Survey classification} & \textbf{Embodied} & \shortstack{\textbf{Post-deployment}\\\textbf{self-improvement}} & \shortstack{\textbf{Body/hardware}\\\textbf{evolution}} & \shortstack{\textbf{Real-time}\\\textbf{safety constraint}} & \shortstack{\textbf{Lifelong / trajectory}\\\textbf{evaluation}} \\
\midrule
Humanoid system reviews~\citep{sheng2025smartbot,cao2024humanoid} & \cmark & \xmark & \xmark & \pmark & \xmark \\
Humanoid perception~\citep{bin2025visualperception}               & \cmark & \xmark & \xmark & \xmark & \xmark \\
Motion planning~\citep{rutili2024motionplanning}                  & \cmark & \xmark & \xmark & \pmark & \xmark \\
VLA models~\citep{ma2024vlasurvey}                                & \cmark & \xmark & \xmark & \xmark & \xmark \\
Self-evolving agents~\citep{gao2026selfevolving,fang2025selfevolving} & \xmark & \cmark & \xmark & \pmark & \cmark \\
Agent evaluation~\citep{yehudai2025agenteval}                     & \xmark & \pmark & \xmark & \xmark & \cmark \\
Continual-learning surveys~\citep{vandeven2025continual}          & \xmark & \pmark & \xmark & \xmark & \cmark \\
World-model surveys~\citep{worldmodelsurvey2025,ding2025worldmodelsurvey} & \pmark & \xmark & \xmark & \xmark & \xmark \\
\textbf{This survey}                                             & \cmark & \cmark & \cmark & \cmark & \cmark \\
\bottomrule
\end{tabular}}
\vspace{-0.1in}
\end{table*}

\subsection{Comparison with Related Surveys and the Position of This Work}
\label{sec:related-surveys}
This survey considers the intersection of two literatures that have so far been studied separately. Specifically, humanoid and VLA surveys classify the capabilities of embodied systems but treat learning as something that happens before deployment. Humanoid system reviews summarize components, milestones, and key technologies, including body design, locomotion control, perception, and manipulation, together with their open challenges~\citep{sheng2025smartbot,cao2024humanoid}. Humanoid perception surveys cover the perception module as a fixed pipeline for state estimation and environmental interaction~\citep{bin2025visualperception}; motion-planning surveys cover the planning and control stack offline~\citep{rutili2024motionplanning}.  VLA surveys provide a broad view of vision-language-action architectures, control policies, and task planners, but generally treat the model as a fixed reasoning-and-acting system without lifelong self-modification~\citep{ma2024vlasurvey}. The second area is self-evolving agents. These surveys study post-deployment improvement in depth~\citep{gao2026selfevolving,fang2025selfevolving}, together with evaluation~\citep{yehudai2025agenteval} and open-ended self-modification~\citep{zhang2026dgm}. However, they mainly focus on software agents. These agents have no physical body, real-time physical safety constraints, or morphology that can change over time. This difference necessitates a unified view of self-evolution that accounts for both the learning process and the physical body of a humanoid.

Table~\ref{tab:surveys} compares this survey with related reviews. To the best of our knowledge, no previous survey treats post-deployment self-improvement as the core principle of an embodied system whose tasks, environment, and body can all change over a multi-year lifetime. We therefore position this survey as the first to bridge these areas. The embodied setting differs in two important ways: (i) the robot’s body can change, and (ii) physical-safety requirements impose hard limits on permissible updates. These issues are not fully addressed in the existing humanoid and self-evolving-agent literature. This motivates our survey, which asks how a deployed humanoid can update its policy, perception, memory, workflow, and body from its own experience while remaining safe throughout its lifetime.

\subsection{Contributions and Organization}
\label{sec:contributions}

The contributions of our work can be summarized as follows:

\begin{itemize}
    \item \textbf{An operational definition and formal framework:} We describe a deployed humanoid by the state tuple: policy, perception, memory, workflow, and body. These states can be updated by an evolution operator on a slow outer loop under a lifelong objective. Then, we give three inclusion criteria that separate self-evolution from ordinary learning.
    \item \textbf{A logical taxonomy with a unified analysis framework:} We organize the field into self-learning, self-adaptation, self-optimization, and self-generation, in increasing order of autonomy. We classify representative methods based on their mechanism, sub-operator, trigger, platform, and update timing. Across these mechanisms, we identify several important trends. Post-deployment changes are increasingly handled by small and reversible modules rather than direct changes to the main model weights. At the same time, the responsibility for verifying these changes is gradually shifting from human supervision to world models. Finally, we show that forgetting is not always harmful; in a continuously changing environment, some forgetting is necessary for the system to adapt.
    \item \textbf{Safety as a property of the evolution operator:} We formulate safe self-evolution as a constrained optimization problem in which every proposed update must remain within an admissible region. We then organize existing safety approaches into a five-layer framework and assess their maturity. This analysis reveals two major gaps. First, existing systems do not explicitly enforce monotonicity to prevent the loss of safety-critical capabilities. Second, current safeguards largely overlook the body operator, leaving an important source of risk unaddressed.
    \item \textbf{Trajectory-level evaluation and benchmarks:} We propose four trajectory-level metrics: adaptation rate, forgetting rate, transient cost, and safety-regression rate to evaluate lifelong self-evolution. Our review shows that standardized evaluation under continuous drift is still missing. To address this gap, we outline a benchmark suite for self-evolution.
\end{itemize}

\begin{table*}[t]
\centering
\caption{Coverage map: each humanoid module realizes a component of the state tuple $s_\tau=(\pi_\tau,\Phi_\tau,\mathcal{M}_\tau,\mathcal{W}_\tau,\mathcal{B}_\tau)$, runs a different approach of learning machinery, and improves from a different signal.}
\label{tab:coverage}
\small
\scalebox{0.80}{
\begin{tabular}{p{0.12\textwidth} p{0.46\textwidth} p{0.31\textwidth} p{0.250\textwidth}}
\toprule
\textbf{Module (state component)} &
\textbf{Representative model approach} &
\textbf{Primary task} &
\textbf{Improves from} \\
\midrule

Perception $\Phi$
&
ViT/CNN encoders; self-supervised representations~\citep{he2022mae,caron2021dino}
&
State estimation and semantic perception
&
Unlabeled sensor streams (self-supervised learning)
\\

Policy/control $\pi$
&
RL policies; VLA action heads, autoregressive~\citep{kim2024openvla} and flow-based~\citep{black2024pi0}
&
Locomotion, whole-body control, and manipulation
&
Rewards; demonstrations
\\

Memory $\mathcal{M}$
&
Skill libraries, replay buffers, and learned world models~\citep{wan2024lotus,hafner2025dreamerv3}
&
Skill reuse; future-state prediction
&
Replay; consolidation
\\

Workflow $\mathcal{W}$
&
Classical motion planning/TAMP~\citep{rutili2024motionplanning}; VLM/LLM planners~\citep{thinkact2025}
&
Task decomposition; long-horizon planning
&
Reflection; execution feedback
\\

Body $\mathcal{B}$
&
Calibration; kinematic/dynamic self-models~\citep{hu2025selfmodel,hu2025simself}
&
Tracking morphology; dynamics
&
Proprioception; self-modeling
\\

\bottomrule
\end{tabular}}
\vspace{-0.1in}
\end{table*}

The remainder of the survey is organized as follows. Section~\ref{sec:foundations} presents the humanoid as a system of coupled modules, gives the operational and formal definition of self-evolution, and introduces the four mechanisms. Sections~\ref{sec:Self-learning} through~\ref{sec:Self-Generation} develop the four mechanisms in turn. Section~\ref{sec:safety} treats safe and uncertainty-aware evolution, Section~\ref{sec:evaluation} and~\ref{sec:challenges} address evaluation and open challenges, respectively, and Section~\ref{sec:conclusion} concludes the survey.

\section{Foundations: Humanoid Modules and a Definition of Self-Evolution}
\label{sec:foundations}

\subsection{The Humanoid as a System of Coupled Modules}
\label{sec:modules}

A humanoid is a complex system of interconnected modules, including perception, planning, control, and actuation, that continuously interact through proprioceptive and exteroceptive feedback. It combines a motion system that provides the required degrees of freedom, a control system that plans and issues commands, and a perception system that closes the loop with the surrounding environment~\citep{bin2025visualperception}. Whole-body stability, generalizable skill learning, and human interaction are among its main technical challenges~\citep{sheng2025smartbot}. This makes humanoids a useful platform for studying self-improvement because different modules use different learning methods, address different tasks, and learn from different types of feedback. Table~\ref{tab:coverage} summarizes these learning methods and maps them to the corresponding state components and tasks.

\subsection{An Operational Definition of Self-Evolving Humanoids}
\label{sec:def}

\subsubsection{The Humanoid as a Non-stationary Process}
The four drivers come from different sources and cause the deployed humanoid to operate in a non-stationary environment that changes over time. The distribution of observations, the dynamics that map actions to outcomes, and even the task space~\citep{zenke2017si} can change with the environment~\citep{dexteroussafe2025}, the robot's body, and the user~\citep{gentlehumanoid2025}. A policy trained on a fixed snapshot of this process may perform well at first but become increasingly suboptimal as these conditions change. Therefore, the goal is not to optimize the humanoid at a single point in time, but to improve the humanoid as it evolves throughout its deployment.

\begin{figure*}[t]
\centering
\includegraphics[width=.92\textwidth]{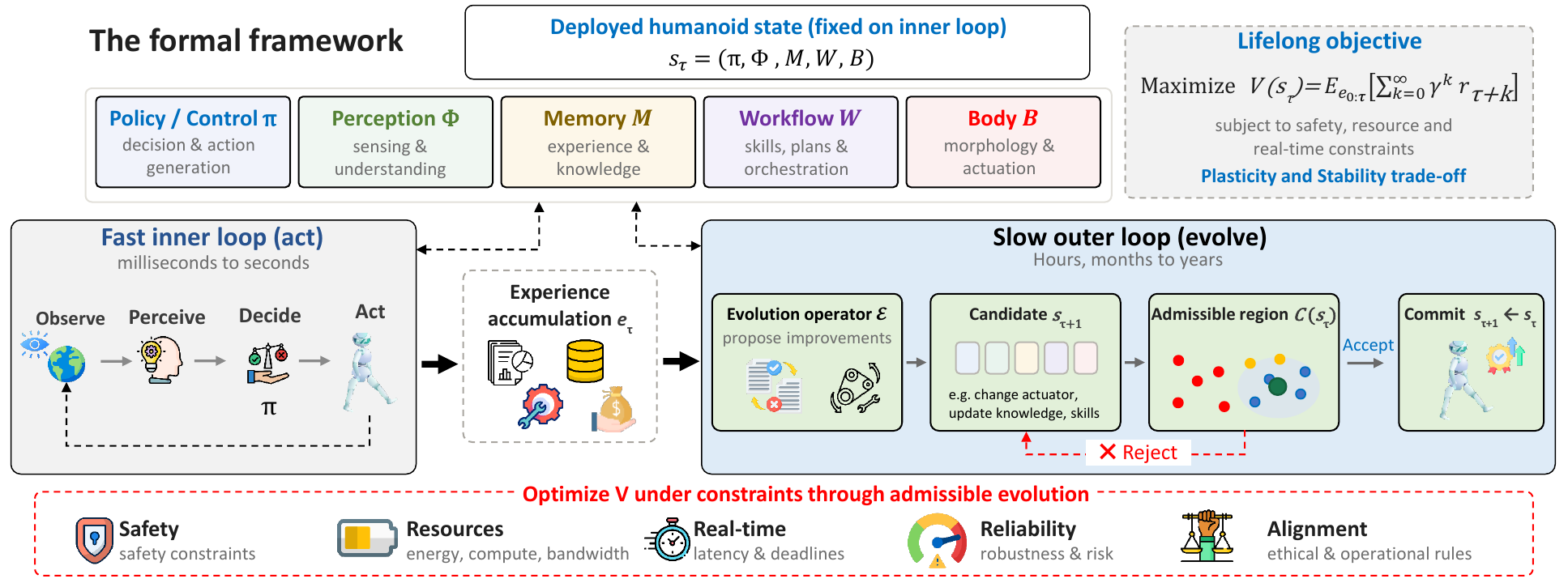}
\caption{The deployed humanoid is the state tuple $s_{\tau}=(\pi,\Phi,\mathcal{M},\mathcal{W},\mathcal{B})$. On the fast inner loop (milliseconds to seconds) the tuple is fixed and the robot acts; on the slow outer loop (hours to months) accumulated experience $e_{\tau}$ drives the evolution operator $\mathcal{E}$, whose candidate $s_{\tau+1}$ must pass the admissible region $\mathcal{C}(s_{\tau})$ before being committed, optimizing the lifelong objective $V$ under the plasticity-stability trade-off.}
\label{fig:framework}
\vspace{-0.1in}
\end{figure*}

\subsubsection{The Operational Definition and Decomposition}
In the following, we define it operationally and establish an explicit scope. Inspired by~\citep{gao2026selfevolving}, we consider a deployed humanoid at evolution time step $\tau$ described by the tuple
\begin{equation}
  s_{\tau} = (\pi_{\tau}, \Phi_{\tau}, \mathcal{M}_{\tau}, \mathcal{W}_{\tau}, \mathcal{B}_{\tau}),
  \label{eq:state}
\end{equation}
where $\pi_{\tau}$ denotes the control and decision policy, $\Phi_{\tau}$ the perception and representation module, including encoders and state estimators, $\mathcal{M}_{\tau}$ the memory, including the skill library, replayed experience, and learned world-model parameters, $\mathcal{W}_{\tau}$ the task composition and planning workflow, and $\mathcal{B}_{\tau}$ the body configuration exposed to the software, including calibration, morphology, sensor, and actuator states. Let $e_{\tau}$ denote the experience collected by the humanoid after deployment. The humanoid self-evolves when an evolution operator $\mathcal{E}$ uses this experience to update its state
\begin{equation}
  s_{\tau+1} = \mathcal{E}(s_{\tau}, e_{\tau}, \theta_{\tau}),
  \qquad
  \theta_{\tau+1} = \mathcal{F}(\theta_{\tau}, e_{\tau}),
  \label{eq:evolve}
\end{equation}
where $\theta_{\tau}$ parameterizes the evolution mechanism itself and $\mathcal{F}$
is a slower meta-update that retunes it. The evolution operator can be decomposed into sub-operators corresponding to the state components:
\begin{equation}
  \mathcal{E} = (\mathcal{E}_{\pi}, \mathcal{E}_{\Phi}, \mathcal{E}_{\mathcal{M}}, \mathcal{E}_{\mathcal{W}}, \mathcal{E}_{\mathcal{B}}),
  \label{eq:operator}
\end{equation}
where each method in this survey can be described by the sub-operator or sub-operators it updates. Each sub-operator can be triggered by different conditions. However, these sub-operators are coupled, so a change in one can also affect the others. For example, a humanoid may update its skill library to perform a new task, which may also require a change in its control policy. The body operator $\mathcal{E}_{\mathcal{B}}$ has no direct counterpart in the disembodied self-evolving-agent setting~\citep{fang2025selfevolving}. A humanoid must also keep its body model up to date as the hardware wears out, is damaged, or is upgraded.

To distinguish self-evolution from ordinary learning, we use three inclusion criteria based on the operational definition of self-evolving agents~\citep{gao2026selfevolving}. An update is considered self-evolution only when it satisfies all three criteria. First, it must be experience-driven. The update should come from the humanoid's own trajectories, self-generated data, or human feedback and should address a specific capability limitation. Second, it must be persistent. The update should produce a lasting change to the system, such as changing the policy or state $s_\tau$, rather than only its temporary behavior. Third, it must be autonomous. The humanoid should have a mechanism for self-initiated exploration, reflection, or reconfiguration, even when pre-collected data are also used. Based on how the update is initiated, we further distinguish active self-evolution, which is initiated by the humanoid itself, from passive self-evolution, which is triggered only by externally provided data and control. Finally, self-evolution differs from other learning paradigms in what it can change. Continual and curriculum learning mainly update model parameters~\citep{julian2025building}, whereas self-evolving systems may also modify non-parametric components, including the skill library $\mathcal{M}$~\citep{meng2025legion}, workflow $\mathcal{W}$~\citep{hierarchicalvlp2025}, and even the physical body $\mathcal{B}$.

\subsubsection{Two-timescale Dynamics and the Lifelong Objective}
\label{sec:twotimescale}

As shown in Figure~\ref{fig:framework}, the system operates on two timescales that should be clearly separated. The \textbf{control step} $t$ indexes the fast inner loop, which runs at the actuation rate. During this loop, the tuple $s_\tau$ remains fixed, and the robot acts according to $a_t=\pi_\tau(x_t)$. The \textbf{evolution step} $\tau$ indexes the slower outer loop, where $\mathcal{E}$ commits an update to $s_\tau$. Self-evolution takes place in this outer loop. Therefore, the goal of a self-evolving humanoid is to improve its cumulative competence throughout deployment:
\begin{equation}
\begin{aligned}
V &= \mathbb{E}\!\left[ \sum_{\tau}\gamma^{\tau}\!\left(
\sum_{t}u_t -\rho_{\mathrm{plas}}D(s_{\tau},s_{\tau+1}) -\rho_{\mathrm{stab}}R(s_{\tau},s_{\tau+1}) \right)
\right] \\ &\text{s.t.}\quad s_{\tau}\in\mathcal{S}_{\mathrm{safe}},\quad \forall\,\tau,
\end{aligned}
\label{eq:lifelong}
\end{equation}
where $\sum_{t} u_{t}$ is the task utility accumulated during the fast loop within one slow step, and $\gamma$ discounts evolution steps rather than control steps. The term $D$ represents the cost of changing the body or policy faster than the changes can be verified, which we call the \emph{plasticity} cost. The term $R$ represents the cost of losing previously acquired competence, which we call the \emph{stability} cost. The weights $\rho_{\mathrm{plas}}$ and $\rho_{\mathrm{stab}}$ control the trade-off between adaptation and retention, reflecting the plasticity-stability tension~\citep{nguyen2026survey}. Finally, $\mathcal{S}_{\mathrm{safe}}$ defines the admissible region and limits which updates are allowed.

\subsection{Four Pillars of Self-Evolution}
\label{sec:pillars}
The remainder of this survey is organized based on four complementary mechanisms of self-evolution that jointly describe how a deployed humanoid can continuously improve over time. We present these mechanisms in increasing order of autonomy. Section~\ref{sec:Self-learning} focuses on self-learning, which enables the humanoid to acquire new knowledge and skills from experience while retaining previously learned capabilities. Section~\ref{sec:Self-Adaptation} then discusses self-adaptation, which allows the humanoid to respond to changes in its environment, body, and interactions with humans. Section~\ref{sec:Self-Optimization} examines self-optimization, which improves existing behaviors through reinforcement learning, world models, planning, and self-reflection. Finally, Section~\ref{sec:Self-Generation} covers self-generation, which allows the humanoid to create new tasks, data, or experiences that can further support learning, adaptation, and optimization. Therefore, these mechanisms form a continuous improvement cycle: learn, adapt, optimize, and generate. They are also closely connected in practice: learning provides new capabilities, adaptation responds to changing conditions, optimization improves existing behaviors, and generation creates new experiences that feed the other three mechanisms. Table~\ref{tab:classification} presents representative works of each mechanism.

\begin{table*}[t]
\centering
\caption{Classification of representative methods by sub-operator(s), trigger (A: active, i.e., self-initiated; P: passive, i.e., externally supplied), demonstrated platform, and update timing.}
\label{tab:classification}
\small
\scalebox{0.81}{
\setlength{\tabcolsep}{4pt}
\begin{tabular}{@{}lcll@{\hspace{2em}}lcll@{}}
\toprule
\textbf{Work} & \textbf{Sub-op.} & \textbf{Trig.} & \textbf{Platform / Timing} &
\textbf{Work} & \textbf{Sub-op.} & \textbf{Trig.} & \textbf{Platform / Timing} \\
\midrule
\multicolumn{4}{@{}l}{\emph{Self-learning} (\S\ref{sec:Self-learning})} &
\multicolumn{4}{l}{\emph{Self-optimization} (\S\ref{sec:Self-Optimization})} \\
LEGION~\citep{meng2025legion} & $\mathcal{E}_{\mathcal{M}},\mathcal{E}_{\pi}$ & P & Manipulator, Epis. &
H2O / OmniH2O / ExBody2~\citep{he2024h2o,he2024omnih2o,exbody2_2024} & $\mathcal{E}_{\pi}$ & P & Humanoid, Off. \\
LOTUS~\citep{wan2024lotus} & $\mathcal{E}_{\mathcal{M}}$ & P & Manipulator, Epis. &
BFM-Zero~\citep{bfmzero2025} & $\mathcal{E}_{\pi}$ & A & Humanoid, Off. \\
LifeLong-RFT~\citep{longlived2026} & $\mathcal{E}_{\pi}$ & P & VLA manip., Epis. &
DreamVLA~\citep{zhang2025dreamvla} & $\mathcal{E}_{\mathcal{M}},\mathcal{E}_{\pi}$ & P & VLA manip., Off. \\
DayDreamer~\citep{wu2022daydreamer} & $\mathcal{E}_{\pi},\mathcal{E}_{\mathcal{M}}$ & A & Quadruped/arms, Online &
WMPO~\citep{wmpo2025} & $\mathcal{E}_{\pi}$  & A & VLA manip., Epis. \\
ResFiT~\citep{residualoffpolicy2025} & $\mathcal{E}_{\pi}$ & P & Humanoid, Online &
REFLECT~\citep{liu2023reflect} & $\mathcal{E}_{\mathcal{W}}$ & A & Manipulator, Epis. \\
RobotTrainsRobot~\citep{robottrainsrobot2025} & $\mathcal{E}_{\pi}$ & P & Humanoid, Online &
Phoenix~\citep{phoenix2025} & $\mathcal{E}_{\pi},\mathcal{E}_{\mathcal{W}}$ & A & Manipulator, Epis. \\
HIL-SERL~\citep{luo2025hitl} & $\mathcal{E}_{\pi}$ & P & Arms, Online &
TT-VLA~\citep{ttvla2026} & $\mathcal{E}_{\pi}$ & A & VLA manip., Online \\
R3M / MVP / VC-1~\citep{nair2022r3m,xiao2022mvp,majumdar2023vc1} & $\mathcal{E}_{\Phi}$ & P & Arms/sim, Off. &
\multicolumn{4}{l}{\emph{Self-generation} (\S\ref{sec:Self-Generation})} \\
PvP~\citep{pvp2025} & $\mathcal{E}_{\Phi}$ & P & Humanoid, Off. &
MimicGen~\citep{mandlekar2023mimicgen} & $\mathcal{E}_{\mathcal{M}}$ & P & Sim manip., Off. \\
\multicolumn{4}{@{}l}{\emph{Self-adaptation} (\S\ref{sec:Self-Adaptation})} &
DexMimicGen~\citep{jiang2025dexmimicgen} & $\mathcal{E}_{\mathcal{M}}$ & P & Humanoid (sim+real), Off. \\
Domain/dyn. rand.~\citep{tobin2017domainrand,peng2018dynamicsrand} & $\mathcal{E}_{\pi}$ & P & Arm/sim, Off. &
DexFlywheel~\citep{dexflywheel2025} & $\mathcal{E}_{\mathcal{M}}$ & A & Dexterous manip., Epis. \\
PolySim~\citep{polysim2025} & $\mathcal{E}_{\pi}$ & P & Humanoid, Off. &
D-CODA~\citep{dcoda2025} & $\mathcal{E}_{\mathcal{M}}$ & P & Bimanual manip., Off. \\
ASAP~\citep{he2025asap} & $\mathcal{E}_{\mathcal{B}},\mathcal{E}_{\pi}$ & P & Humanoid, Epis. &
DreamGen~\citep{dreamgen2025} & $\mathcal{E}_{\mathcal{M}}$ & A & Humanoid VLA, Off. \\
MOSAIC~\citep{mosaic2026} & $\mathcal{E}_{\pi}$ & P & Humanoid, Epis. &
Eureka / EurekaVerse~\citep{ma2024eureka,eurekaverse2024} & $\mathcal{E}_{\mathcal{W}},\theta$ & A & Sim/quadruped, Epis. \\
TTT parkour~\citep{tttparkour2026} & $\mathcal{E}_{\pi},\mathcal{E}_{\Phi}$ & P & Humanoid, Online &
RoboGen / GenSim~\citep{wang2023robogen,wang2023gensim} & $\mathcal{E}_{\mathcal{W}},\mathcal{E}_{\mathcal{M}}$ & A & Sim, Off. \\
EvolveVLA~\citep{evolvevla2025} & $\mathcal{E}_{\pi}$ & A & VLA manip., Online &
& & & \\
Egocentric self-model~\citep{hu2025selfmodel} & $\mathcal{E}_{\mathcal{B}}$ & A & Legged robot, Online &
& & & \\
Self-simulation~\citep{hu2025simself} & $\mathcal{E}_{\mathcal{B}},\mathcal{E}_{\mathcal{M}}$ & A & Robot self-models, Epis. &
& & & \\
Active preference GP~\citep{biyik2024preference} & $\theta$ (objective) & A & Arm/sim, Epis. &
& & & \\
\bottomrule
\end{tabular}}
\vspace{-0.1in}
\end{table*}

\section{Self-Learning in Humanoids}
\label{sec:Self-learning}
The first mechanism focuses on a basic requirement for a deployed humanoid: continuously gaining new capabilities from its changing stream of experience without losing previously learned knowledge. In the notation of Section~\ref{sec:def}, this corresponds to the operators $\mathcal{E}_{\pi}$, $\mathcal{E}_{\Phi}$, and $\mathcal{E}_{\mathcal{M}}$ acting on the live experience stream $e_{\tau}$ under distributional drift. The central challenge is the plasticity-stability trade-off: adapting too slowly leaves the humanoid unable to keep up with a changing environment, while adapting too aggressively can cause it to forget previously learned knowledge and skills.

\subsection{Continual Learning in Humanoids}
\label{sec:Continual}
\begin{figure*}[t]
\centering
\includegraphics[width=0.8\textwidth]{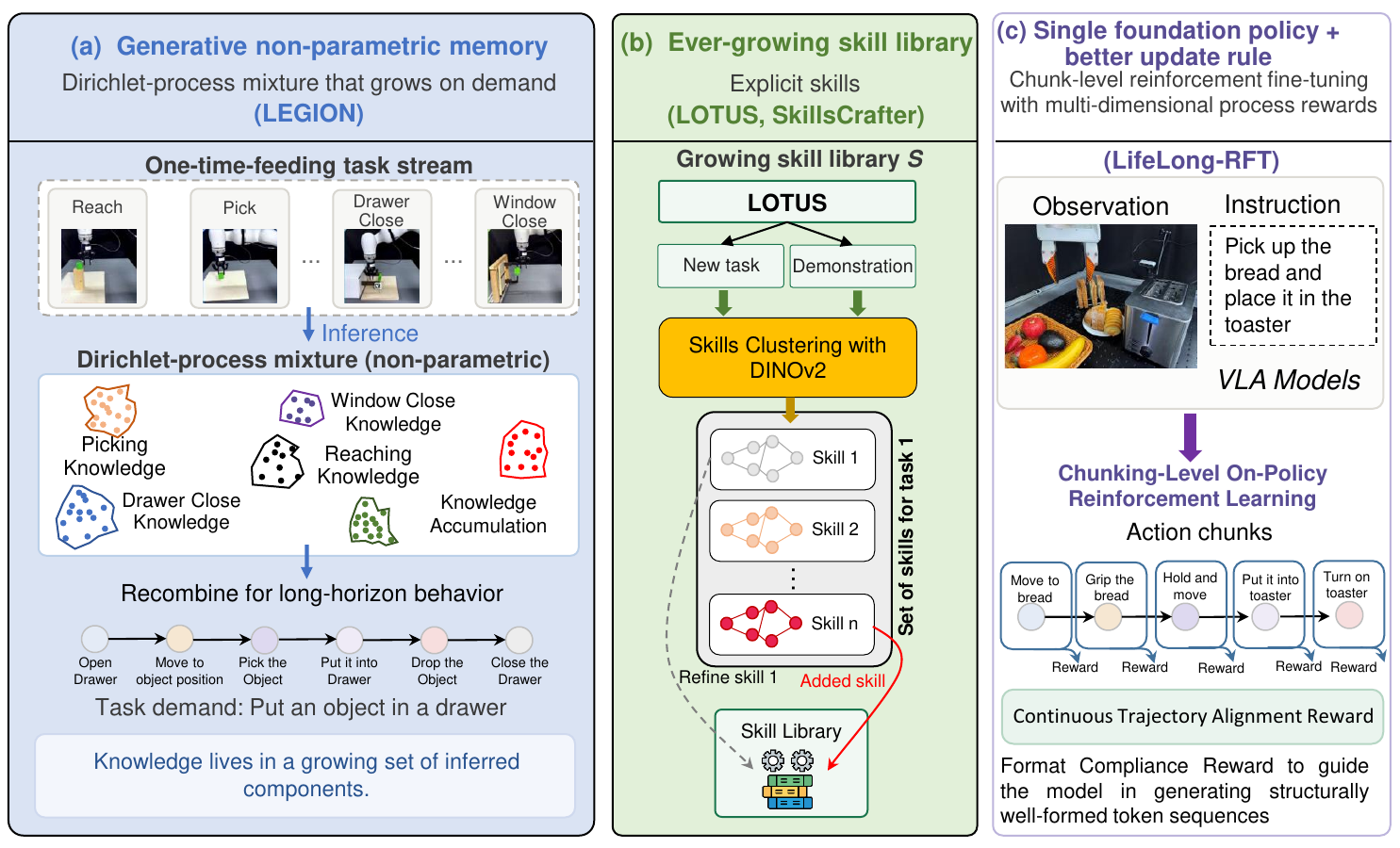}
\caption{Three architectural address to catastrophic forgetting, distinguished by where accumulated knowledge lives: (a) A generative non-parametric memory~\citep{meng2025legion}, (b) An explicit, ever-growing skill library with a routing meta-controller~\citep{wan2024lotus}, (c) A single foundation policy with a revised update rule, in which chunk-level reinforcement fine-tuning with multidimensional process rewards replaces forgetting-prone supervised fine-tuning~\citep{longlived2026}.}
\label{fig:forgetting}
\vspace{-0.15in}
\end{figure*}

Self-learning addresses a challenge that conventional policy learning largely avoids: enabling a humanoid to acquire new skills from its own experience without overwriting previously learned capabilities. Current humanoid controllers typically avoid this problem by training whole-body or manipulation policies in simulation or from demonstrations and then freezing them after deployment. As a result, neither further learning nor forgetting occurs~\citep{kim2024openvla,black2024pi0}. In contrast, continual self-learning operates on a live stream of sensorimotor experience, represented as $e_{\tau}$, and updates the policy as its observed competence changes over time. The central challenge is therefore catastrophic forgetting. Existing embodied-learning approaches offer three distinct ways to address it, differing mainly in where and how previously acquired knowledge is retained, as illustrated in Figure~\ref{fig:forgetting}.

The first approach stores knowledge in a generative, non-parametric memory. LEGION~\citep{meng2025legion} combines an off-policy reinforcement learner with a knowledge space based on a Dirichlet process mixture model. Tasks arrive sequentially, and each inferred task representation is either assigned to an existing mixture component or used to create a new one. Specifically, a pre-trained language model produces a language embedding of the task instruction, which is combined with the observed state to improve task identification. Because the number of mixture components grows with the observed data rather than being fixed, learning a new task does not overwrite representations of earlier tasks. In the reported experiments, the robot achieves an average forgetting score of 0.0 across ten manipulation tasks, with an average success rate of 0.84~\citep{meng2025legion}. More importantly, the robot can complete long-horizon real-world tasks by chaining knowledge acquired from separate single-task episodes.

The second approach stores knowledge in an explicit, continuously growing skill library. LOTUS~\citep{wan2024lotus} performs continual skill discovery from unsegmented human demonstrations using an open-world vision model. Each demonstration is first divided into temporal segments through agglomerative clustering of DINOv2 features, which provide semantically consistent representations under the changing data distribution of lifelong learning. When a new task arrives, its segments are compared with existing skill partitions using a Silhouette criterion. Segments above a similarity threshold are merged into an existing partition and used to refine that skill, while segments below the threshold create a new partition and train a new skill. Each skill is represented as a goal-conditioned visuomotor policy operating directly on raw images, while a transformer-based meta-controller selects the skill and its subgoal at each step. On a 50-task real-robot benchmark, this design improves forward transfer by 11\% over an experience-replay baseline while reducing negative backward transfer. In simulation, it also reduces negative backward transfer on LIBERO-OBJECT from 24.0 to 11.0.

SkillsCrafter~\citep{skillscrafter2026} extends the library idea to the parameter space of a language-conditioned manipulation policy. Each skill is learned by applying low-rank adaptation to a frozen backbone. The two adapter factors have different roles: one captures skill-shared knowledge, while the other is constrained to be orthogonal to the factors of previous skills. This prevents new skills from overwriting parameters learned for earlier skills. To index the skill library, the method applies singular value decomposition to the instructions of each skill and stores the resulting semantic subspace projections. A skill-specialization aggregation module then projects a new instruction onto each stored subspace. Across 18 manipulation skills, the method achieves an average success rate of 52.0\% with 16.0\% forgetting, compared with 50.0\% success and 20.8\% forgetting for the strongest adapter-combination baseline.

The third approach stores knowledge in the weights of a single foundation policy and instead changes how the policy is updated. Supervised fine-tuning of a VLA model requires substantial data and can lead to forgetting. In contrast, on-policy reinforcement learning updates the policy using its own samples and has been shown to be more resistant to forgetting. LifeLong-RFT~\citep{longlived2026} uses this property to continually fine-tune the model with chunk-level on-policy reinforcement learning. It uses a multi-dimensional process reward that evaluates each intermediate action chunk of the humanoid based on three factors: discrete token consistency, continuous trajectory alignment with the reference trajectory, and output-format compliance. Since the reward is computed from expert demonstrations, the method does not require a simulator, reward model, or physical interaction during fine-tuning. In the LIBERO benchmark~\citep{liu2023libero}, the robot trained by the proposed method improves the area under the success-rate curve by 15.1 to 35.9 points over supervised fine-tuning of the same backbone, while using only 10 demonstrations for each new task. These results suggest that forgetting depends not only on how knowledge is stored, but also on how the model is updated.

All three approaches implement the cognitive-evolution operator $\mathcal{E}_{\mathcal{M}}$ while addressing the problem of forgetting. They also adopt common continual-learning techniques, including regularization~\citep{kirkpatrick2017ewc,zenke2017si,aljundi2018mas}, replay~\citep{lopezpaz2017gem}, and parameter isolation~\citep{rusu2016progressive}, as discussed in~\citep{vandeven2025continual}. These techniques are applied to embodied systems and evaluated on lifelong-transfer benchmarks such as LIBERO. A clear trend emerges across these approaches. Knowledge is increasingly stored in structured forms that allow new skills to be added without overwriting existing ones. Examples include mixture components, skill libraries, and orthogonal adapters. In contrast, foundation-policy approaches keep knowledge in a single model and rely on the update rule to preserve previously learned capabilities. A similar pattern appears in disembodied self-evolving agents, which build textual libraries of skills and insights~\citep{wang2024voyager,zhao2024expel}, while robots build sensorimotor libraries. However, the embodied setting has an important difference: some forgetting is necessary. Retired hardware and outdated calibration may need to be removed rather than preserved, which we discuss later.

\subsection{Online Learning in Humanoids}
\label{sec:Online}

Continual learning and online learning address different questions. Continual learning focuses on what a policy retains as it learns across tasks, whereas online learning focuses on how quickly a policy improves through continuous interaction with the environment. The feasibility of online learning on physical robots has been demonstrated by DayDreamer~\citep{wu2022daydreamer}. The Dreamer-based world-model agent was deployed on four robots and learned directly in the real world without a simulator. An A1 quadruped learned to roll over, stand, and walk from scratch in about one hour without manual resets. After the initial training, the robot adapted to new perturbations within ten minutes. It learned to withstand light pushes and to roll over and stand again after hard pushes. Two robotic arms also learned visual pick-and-place tasks from sparse rewards and achieved performance close to human teleoperation. Notably, the same hyperparameter setting was used across all four platforms.
Two lessons emerge from this study. First, the world model makes on-hardware learning practical by converting each real transition into many imagined ones. This supports the argument in Section~\ref{sec:World} that the world model provides the foundation for self-improvement. Second, the push-recovery experiment provides an early hardware demonstration of the perturb-degrade-re-adapt process. Section~\ref{sec:eval-metrics} formalizes this process through the adaptation rate.

Building on these ideas, subsequent systems have explored ways to make real-world adaptation more practical through additional guidance, constraints, and environmental support. HIL-SERL~\citep{luo2025hitl} incorporates human guidance through three components. A pretrained visual backbone stabilizes training, a sample-efficient off-policy learner is initialized with 20 to 30 demonstrations, and a human operator provides corrections that are stored in both the demonstration and on-policy buffers. The learner samples equally from the two buffers. Across dexterous robotic tasks, the method achieves 100\% success within 1 to 2.5 hours of real-world training and executes the tasks 1.8$\times$ faster than imitation-learning baselines using the same amount of human data. The intervention rate decreases to zero during training. Without human corrections, success drops to 49\% even with ten times more demonstrations. This result indicates that human assistance serves as a temporary scaffold for learning.

\begin{figure*}[t]
\centering
\includegraphics[width=0.95\textwidth]{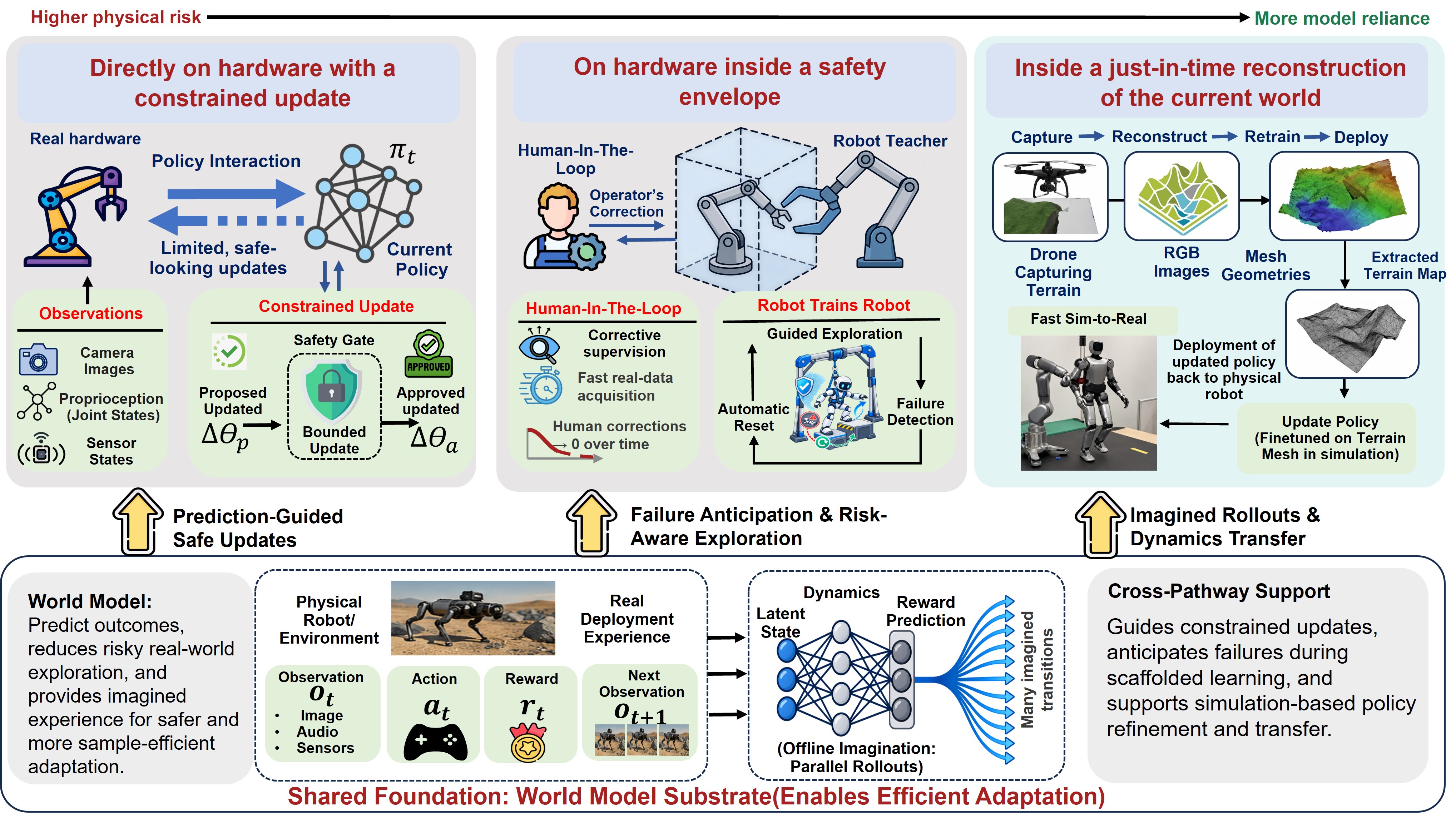}
\caption{Three loci of online improvement ordered by decreasing physical risk and increasing reliance on learned models: directly on hardware with a constrained update ResFiT~\citep{ residualoffpolicy2025}, on hardware inside a safety envelope Robot-Trains-Robot~\citep{robottrainsrobot2025}, and inside a just-in-time reconstruction of the current world TTT-Parkour~\citep{tttparkour2026}. DayDreamer’s world model provides the key mechanism by turning each real interaction into many imagined ones.}
\label{fig:online-loci}
\vspace{-0.1in}
\end{figure*}

ResFiT~\citep{residualoffpolicy2025} removes the human guidance but instead constrains the search space. It uses a two-phase training process. In the first phase, a behavior-cloned base policy is trained. The base policy is then kept frozen, while off-policy RL learns only a lightweight per-step residual correction from sparse binary rewards in the second phase. This avoids optimizing a large policy during online learning and improves sample efficiency. It also enables the first successful real-world RL training on a humanoid with dexterous hands, using a 29-degree-of-freedom platform for bimanual handover.
Robot-Trains-Robot~\citep{robottrainsrobot2025} takes a different approach by using one robot to support another. A robot-arm teacher with force-torque sensing physically assists the humanoid student. This setup provides a safer environment for exploration, a learning curriculum, and informative rewards. The work also introduces a sim-to-real fine-tuning algorithm with three stages: dynamic-aware training, optimization of a universal latent vector for robust initialization, and final fine-tuning. This design improves sample efficiency.
TTT-Parkour~\citep{tttparkour2026} moves adaptation from the real world to simulation. The humanoid first captures the new terrain using RGB-D sensors. A feed-forward pipeline then reconstructs the terrain as a collision-accurate mesh. The policy is fine-tuned in simulation using the reconstructed terrain. The complete capture-reconstruct-retrain process takes less than ten minutes. This allows the humanoid to adapt to new terrains and turn failures on wedges, stakes, and narrow beams into successful executions.

In summary, these systems show three ways to achieve online improvement, as illustrated in Figure~\ref{fig:online-loci}. The first updates the policy directly on the physical robot while keeping the search space small and safe, as in ResFiT. The second also learns on hardware but uses a safety framework provided by a human operator or a robot teacher, as in HIL-SERL and Robot-Trains-Robot. The third avoids most hardware exploration by adapting in a simulation reconstructed from the current environment of the robot, as in TTT-Parkour. These approaches form a progression from direct on-hardware adaptation to increasingly model-based adaptation. Despite these differences, they share an important constraint: every policy update is either performed on the physical robot or transferred to it immediately. Unsafe actions can therefore cause physical damage, so exploration must be carefully controlled, unlike in disembodied settings~\citep{acikgoz2025ttsi}, which motivates the study of online Robust RL~\citep{wang2021online}.

\subsection{Self-Supervised Learning in Humanoids}
\label{sec:Self-supervised}

A humanoid continuously collects experience without labels, and self-supervised learning (SSL) provides a way to use this experience. SSL learns representations $\Phi$ and predictive structure directly from the raw data without human annotation. The learning signal therefore comes from the data itself rather than from an external teacher. The value of this approach is demonstrated by pre-training methods for robotic manipulation. Reusable Representation for Robotic Manipulation~\citep{nair2022r3m} pre-trains on the Ego4D corpus of human videos using time-contrastive learning, video-language alignment, and a sparsity penalty. The resulting frozen encoder improves downstream task success by more than 20\% compared with training from scratch and by more than 10\% compared with CLIP-style representations. It also enables a real robot to learn tasks in a cluttered apartment from only 20 demonstrations. Masked visual pre-training~\citep{xiao2022mvp} shows that a purely reconstructive objective can scale further. A 307M-parameter vision transformer is trained with masked autoencoding on 4.5 million in-the-wild and egocentric frames and then frozen. On real-robot tasks, it outperforms CLIP by up to 75\% and both supervised ImageNet pre-training and training from scratch by up to 81\%. These results suggest that predicting missing visual content can capture control-relevant structure that semantic supervision may miss.

The VC-1 study~\citep{majumdar2023vc1} systematically evaluates this SSL on CortexBench, a suite of 17 tasks covering locomotion, navigation, dexterous manipulation, and mobile manipulation. Three main findings emerge. First, no pre-trained visual representation performs best across all tasks. Second, increasing the size and diversity of the pre-training data generally improves performance, but not for every task. Third, VC-1 matches or exceeds the best reported results only after task-specific or domain-specific \emph{adaptation}. The last finding is particularly important for this survey. There is still no evidence that a single frozen visual encoder is sufficient for all embodied tasks. The perception operator $\mathcal{E}_{\Phi}$ must therefore continue to adapt after deployment, and SSL provides a label-free signal for this adaptation. The encoders discussed above build on masked autoencoding, self-distillation, and contrastive learning objectives~\citep{he2022mae,caron2021dino,chen2020simclr}.

A humanoid-specific version of this idea uses the robot's own body as the source of supervision. PvP~\citep{pvp2025} conducts contrastive learning between the proprioceptive state from the real robot and the privileged simulator state from the simulator. This contrastive alignment requires no hand-crafted augmentations and allows the real robot to learn a compact latent that captures information available in the simulator. On the LimX Oli humanoid, it improves both sample efficiency and final performance in whole-body control for velocity tracking and motion imitation. A similar idea is used in contrastive knowledge distillation for sim-to-real locomotion~\citep{contrastivesim2real2025}. The actor learns to encode proprioceptive information that captures the terrain context available to the critic during simulation. The deployed policy can then infer ground geometry from the robot’s motion and body dynamics without exteroceptive sensing. The method transfers zero-shot to a full-size humanoid, which can climb steps up to 30 cm high and slopes up to 26.5$^{\circ}$. From the perspective of self-evolving agents, SSL is the perceptual counterpart of self-rewarding and self-play, where the training signal is generated internally rather than provided by an external annotator~\citep{yuan2024selfrewarding,zhao2025absolutezero}. The embodied setting has an additional advantage: proprioception is always available from the robot itself and can provide supervision across different environments and tasks.

\vspace{-0.1in}

\subsection{Section Summary and Takeaways}
Self-learning in humanoids can be viewed through three related questions: what the robot should retain when it learns new tasks, how quickly it can improve from ongoing experience, and what it can learn from unlabeled experience. Continual learning addresses the first question by preserving existing skills through structured memories~\citep{meng2025legion}, skill libraries~\citep{wan2024lotus}, or update rules that reduce interference with previously learned knowledge~\citep{longlived2026}. Online learning addresses the second question by enabling real-world adaptation through constrained updates, human or robot assistance, or simulations reconstructed from the robot's current environment~\citep{tttparkour2026}. Self-supervised learning provides a third route: the humanoid learns useful representations and predictive signals directly from sensory and proprioceptive experience, without human annotations~\citep{pvp2025}.

In summary, these studies show that the main challenge is not the lack of mechanisms for post-deployment learning. Such mechanisms are already emerging. The deeper challenge is selective adaptation. A deployed humanoid must determine when its current knowledge is still valid and when it needs to be updated~\citep{rapt2026}. This is difficult because changes are rarely explicit. A new payload, actuator wear, sensor drift, or a change in the environment may first appear as degraded task performance or unexpected body responses~\citep{hu2025selfmodel,splitadapter2026}. This leads to an important difference from self-evolving software agents. Forgetting is not always a failure for a physical robot. Some knowledge may need to be deliberately discarded. Therefore, the plasticity-stability trade-off should not be treated as a fixed constraint. A self-evolving humanoid must manage both plasticity and stability as its body, environment, and tasks change. The system needs to decide how much knowledge to adapt and how much to preserve. This makes the $\rho_{\mathrm{plas}}$ and $\rho_{\mathrm{stab}}$ terms in Eq.~\eqref{eq:lifelong} more than theoretical coefficients. They represent an ongoing decision about whether new experience should modify, reinforce, or replace existing knowledge.

\vspace{-0.1in}

\section{Self-Adaptation in Humanoids}
\label{sec:Self-Adaptation}

The second mechanism concerns adaptation to changes that the humanoid does not initiate. These changes may come from the environment, such as a domain shift, from the robot's own body, such as morphology changes or hardware degradation, or from the humans it serves, such as changes in mood or requirements. While self-learning improves from the experience it receives, self-adaptation is triggered by a detected or observed change in the environment or the robot. In this setting, the embodiment-only operator $\mathcal{E}_{\mathcal{B}}$ becomes central.

\subsection{Domain Adaptation in Humanoids}
\label{sec:Domainadaptation}
The most studied shift is the gap between the distribution used to train a humanoid and the distribution encountered after deployment, which leads to the sim-to-real and real-to-real problems. Two strategies are commonly used, and they differ in when the adaptation is performed.

\subsubsection{The Absorb-a-distribution Approach}

The first strategy, which we term \textbf{absorb-a-distribution}, addresses the shift before deployment. A single policy is trained across a wide range of conditions so that the conditions encountered after deployment are already covered by the training distribution. This approach originates from randomizing visual appearance~\citep{tobin2017domainrand} and system dynamics~\citep{peng2018dynamicsrand}. Recent humanoid studies have focused more on selecting the right distribution to absorb.
PolySim~\citep{polysim2025} identifies \emph{simulator inductive bias} as one source of the remaining sim-to-real gap. Different simulators make different assumptions about the system dynamics, so training with a single simulator can limit the diversity of the learned policy. PolySim addresses this issue by training whole-body controllers across multiple heterogeneous simulators in parallel within a single run. Randomizing the dynamics across simulators reduces the bias of any single simulator and improves cross-simulator execution success by 52.8\% over a baseline trained only in IsaacSim. The resulting policy also transfers to a real Unitree G1 robot without additional training. This approach can further be combined with explicit physics alignment between simulation and hardware~\citep{he2025asap} and teacher-student distillation using privileged information~\citep{he2024omnih2o}.

DoorMan~\citep{pixel2action2025} applies the absorb strategy to the perception stack. It uses a teacher-student bootstrap pipeline with staged-reset exploration and GRPO-based sim-to-real reinforcement fine-tuning to learn an RGB-only humanoid loco-manipulation policy. The policy is trained entirely on randomized photorealistic synthetic data and can open a variety of real doors without prior real-world experience. It also reduces completion time by up to 31.7\% compared with human teleoperators. The cost of this strategy is also decreasing. A stabilized off-policy method can train sim-to-real humanoid locomotion under strong randomization in 15 minutes on a single consumer GPU~\citep{sim2real15min2025}. Retraining the absorber can therefore become part of the main evolution loop rather than a long offline process.

\vspace{-0.1in}

\subsubsection{The Adjust-online Approach}

The second strategy corrects the remaining gap after deployment. MOSAIC~\citep{mosaic2026} shows how to adapt a policy while preserving its generality. Its generalist humanoid motion tracker degrades under interface-specific teleoperation errors. To address this issue, an interface-specific policy is trained with minimal data and then distilled back into the generalist through an additive residual module. This module outperforms both naive fine-tuning and continual-learning updates of the base policy. A similar post-deployment approach is used for VLA policies, where test-time adaptation uses feedback from the environment to improve the policy~\citep{evolvevla2025,acikgoz2025ttsi}. The two strategies are complementary. Randomization increases the policy's tolerance before deployment, while online adaptation updates the policy after the environment changes. However, neither strategy specifies when adaptation should be triggered. Deployment-time monitoring can provide this signal. RAPT~\citep{rapt2026} treats domain shift as an observable signal that can be monitored together with a pre-trained control policy. It models the nominal spatio-temporal behavior of the system and detects deviations during execution. With an episode-level false-positive rate of 0.5\%, it improves out-of-distribution detection by 37\% over the strongest baseline in simulation and by 12.5\% on hardware.

Such monitoring makes self-adaptation self-directed by detecting deviations from expected behavior and triggering an update. The continuous mismatch signal also estimates the magnitude of the distribution shift, supporting an uncertainty-adaptive trust region. Online adjustment is the embodied counterpart of test-time self-improvement in software agents~\citep{acikgoz2025ttsi}. The key difference is that software agents are often informed when they fail, while a humanoid must detect failures from its own observations.

\vspace{-0.1in}

\subsection{Morphology Adaptation in Humanoids}
\label{sec:MorphologyAdapt}

A robot cannot step outside its own body and observe it in the same way that it observes the environment. The body is both the component that changes and the source of the measurements used to detect those changes. Actuator wear, joint backlash, sensor drift, payload changes, or a repaired or replaced hand can all change the body without providing an explicit signal. As a result, the body model $\mathcal{B}$ used by the controller may no longer accurately describe the robot. Two complementary lines of work address this problem, and they place the operator $\mathcal{E}_{\mathcal{B}}$ at different points in the system.

The first line of work treats the body model as an object of \emph{inference}. The robot learns and maintains an explicit self-model from its own sensor data. Hu \emph{et al.}~\citep{hu2025selfmodel} show that a task-agnostic dynamic model of a 12-DoF legged robot can be learned in a self-supervised manner from a single first-person camera, without prior knowledge of its morphology, kinematics, or task. The model can be continuously compared with incoming observations to detect changes, such as a damaged component, and adapt the robot's behavior accordingly. The same approach also generalizes across different robot configurations. A related approach uses an external camera to observe the robot. Hu \emph{et al.}~\citep{hu2025simself} use external visual observations to model the robot's morphology, kinematics, and motor control.

Both works discussed above assume that the robot already knows which body in the camera view is its own. Chen \emph{et al.}~\citep{selfother2026} address this problem by using the correspondence between commanded and observed motion. The body that moves in response to the robot's motion command is identified as its own body. This is also consistent with the cue used in developmental psychology to explain self-recognition in infants. Based on this cue, a humanoid robot can identify its own body among several other bodies using only proprioception and vision, without labels or a kinematic model. The robot then uses this self-recognition to construct a three-dimensional model of its body for reaching, obstacle-aware planning, and human motion imitation. The resulting self-model is therefore not a fixed, factory-calibrated description. Instead, the robot can verify and update it over time, which is important when the body changes.

The second line of work accounts for differences between robot bodies within the controller and limits post-deployment changes to small and controlled adaptations. Two studies focus on morphology-level generality. A unified CrossQ policy for fall recovery~\citep{getupmorph2025} is trained across seven humanoid robots and evaluated zero-shot on unseen robots, achieving up to $86\pm7\%$ success. H-Zero~\citep{hzero2025} follows a similar goal for locomotion. It uses transformation layers at the policy input and output to standardize control semantics, together with randomized physical parameters, varied policy observations, and exploratory learning strategies.
SplitAdapter~\citep{splitadapter2026} focuses on a single robot adapting to changes in payload and pickup height. The authors argue that history-based adapters combine object-induced load changes and robot-side dynamics mismatch into a single latent, which reduces whole-body control performance under heavy loads. To address this issue, they freeze a pretrained AMP-based loco-manipulation policy and add two context branches. One estimates the object mass and loaded state, while the other learns a residual-dynamics latent using split world-model objectives and cross-adversarial regularization. The two branches modulate the early and late layers of the frozen policy through hierarchical FiLM, respectively. The results show that lift-up success is high for all methods, but full-task success shows a clear difference. In the out-of-distribution 6~kg setting, SplitAdapter achieves 93.3\% success, compared with 60\% for the unified-latent adapter. Notably, the policy weights remain unchanged at deployment. The adaptation is therefore performed through inference over a frozen policy rather than through continued learning.
FAST~\citep{generalwbc2026} takes a similar approach by freezing the pretrained controller and adding a small residual policy that learns only the correction needed for a new motion. Two constraints limit this correction. The first keeps the internal features well conditioned, while the second keeps the new behavior close to the behavior learned by the base policy. The robot can therefore learn new motions from text prompts or low-quality videos without forgetting previous skills.

In our notation, morphology adaptation is represented by the operator $\mathcal{E}_{\mathcal{B}}$, which updates the body configuration. This update is closely coupled with $\Phi$, since a new sensor changes what the robot can observe, and with $\pi$, since a change in the body changes what the robot needs to control. A common design pattern appears across these studies. The residual modules in MOSAIC, the orthogonal adapters in SkillsCrafter, and the structured adapters in SplitAdapter and FAST protect the pretrained generalist and limit changes to a constrained part of the model. This idea also anticipates the trust-region layer. The difference from self-modifying software agents is important. Software agents can modify their own architecture or code~\citep{hu2025adas,zhang2026dgm}, whereas a humanoid must also adapt its \emph{body} model.

\begin{figure*}[t]
\centering
\includegraphics[width=0.90\textwidth]{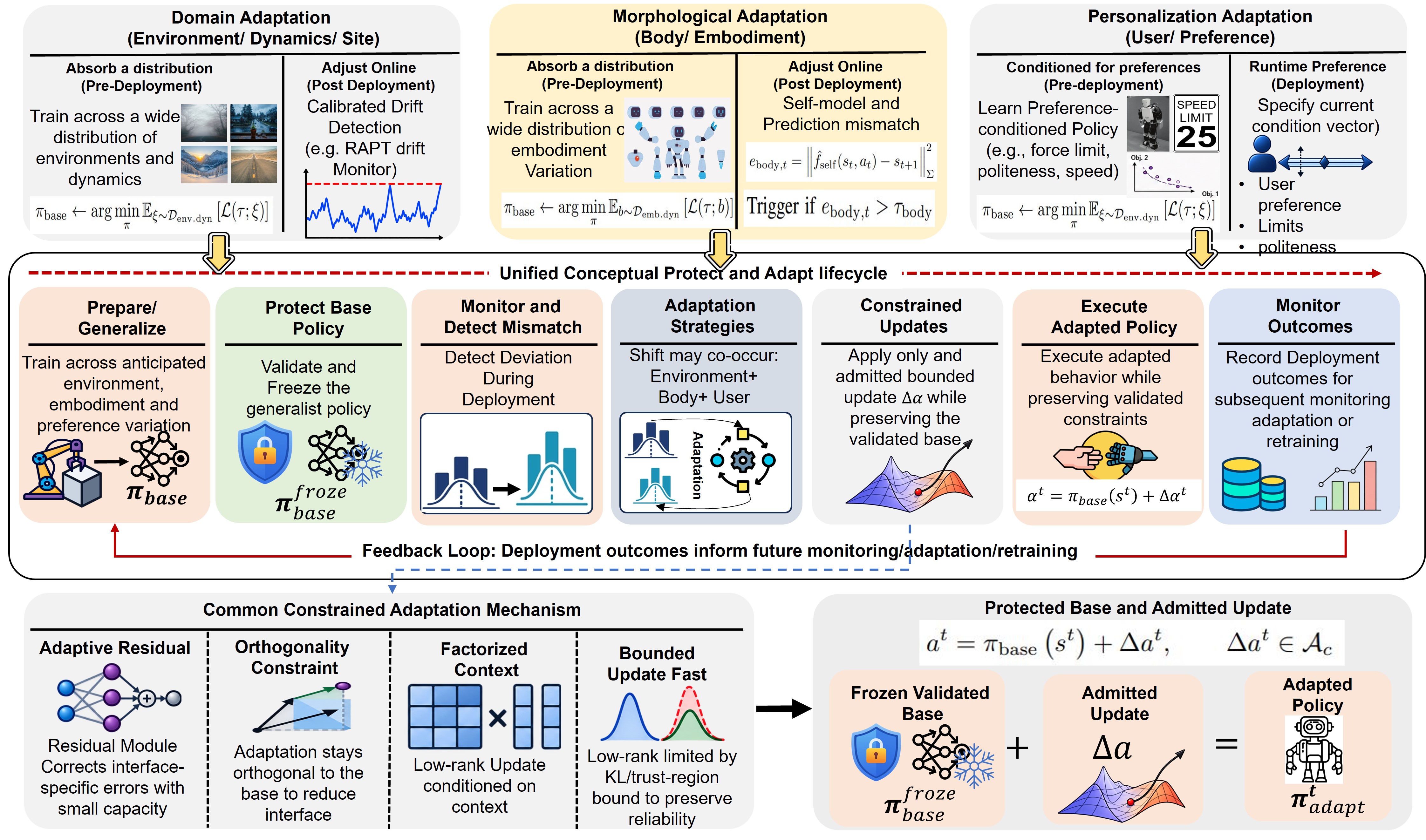}
\caption{
Conceptual framework for constrained humanoid self-adaptation across domain, morphological, and personalization shifts. Domain adaptation combines pre-deployment distribution absorption~\citep{polysim2025} with deployment-time mismatch detection and online adjustment~\citep{mosaic2026,rapt2026}. Morphological adaptation similarly couples embodiment generalization and self-modeling~\citep{hu2025selfmodel,getupmorph2025} with constrained post-deployment corrections~\citep{splitadapter2026,generalwbc2026}, while personalization learns or exposes user preferences as conditioning variables or explicit interaction constraints~\citep{biyik2024preference,prefmulti2025}. These complementary mechanisms are synthesized into a unified protect-and-adapt lifecycle in which a validated base policy is preserved, only admitted bounded updates are executed, and deployment outcomes feed subsequent monitoring, adaptation, or retraining.}
\vspace{-0.1in}
\label{fig:adapt-pattern}
\end{figure*}

\subsection{Personalization in Humanoids}
\label{sec:personalization}

The third shift concerns the individual human that the humanoid serves. Preference-based methods address this problem by adapting the robot's behavior to user preferences without requiring a hand-designed reward. The Gaussian-process method of Bıyık \emph{et al.}~\citep{biyik2024preference} provides a statistical basis for this approach. It models the reward non-parametrically over the trajectory space, allowing it to capture preferences that may not fit a fixed feature set. The robot also selects its own queries by asking for the comparison that is expected to provide the most information. This allows a user-specific reward to be learned from a small number of pairwise preference judgments, which provides a simple form of feedback for non-expert users.

Humanoid-specific work shows that a learned or specified preference does not need to be stored in the policy weights. Preference-conditioned reinforcement learning~\citep{prefmulti2025} addresses the trade-off between command tracking and force compliance. The humanoid must follow velocity commands while also yielding to sustained external forces. The two objectives are formulated as a multi-objective optimization problem, and a single policy is conditioned on the preference weights between them. This allows the humanoid to balance user preferences against current environmental conditions at deployment time, without additional training or changes to the policy architecture.
GentleHumanoid~\citep{gentlehumanoid2025} applies the same idea to physical compliance rather than command tracking. It adds impedance control to a whole-body motion tracking policy by attaching a virtual spring to each upper-body link, anchored at the point of first contact when the robot presses against something and at a pose drawn from human motion data when a person pushes or pulls the arm. Because the anchors are based on complete postures, the shoulder, elbow, and wrist respond together rather than independently. The policy also receives the safe force limit as an input (randomized [5-15]~N) during training. This allows one policy to handle both gentle and supportive interactions: 5~N is gentle enough to hold a balloon, while 15~N is strong enough to help someone stand up. On a Unitree G1, the arm yields at around 10~N, whereas rigid tracking baselines require 24.6~N and 51.1~N to be moved.

These systems are the embodied form of learning from human feedback~\citep{christiano2017deeprlhf,ouyang2022instructgpt}, but embodiment adds physical force to the preference being learned. A user preference therefore affects not only what the humanoid does, but also how much force it applies. An incorrect preference can thus become a safety issue rather than only a quality issue. The second lesson concerns system design. When preferences are provided as conditioning inputs or explicit force limits, they can be changed without retraining the policy. Personalization is therefore efficient to apply, easy to reverse, and easy to inspect. These properties are desirable for any update and motivate the alignment gate introduced in Section~\ref{sec:safety}.

\subsection{Section Summary and Takeaways}

The three shifts discussed in the previous subsections share a common structure: part of the adaptation is performed before deployment, while the remaining adaptation is handled after deployment. Training across randomized environments, different robot bodies, and a range of user preferences expands what a single policy can handle~\citep{polysim2025}. Conditions outside this range are then addressed after deployment through a self-model or a small adaptation module attached to a frozen generalist~\citep{mosaic2026}. Embodiment differs from the software setting in one important aspect: the robot itself can change. Domain adaptation has a close counterpart in test-time adaptation, and personalization has a counterpart in learning from human feedback. A software state has no such counterpart, because no line of code wears down or breaks over time with use.

Two gaps remain. The first is the trigger for adaptation. A humanoid can detect changes in its environment using a calibrated monitor, and its self-model can be continuously checked against its observations. However, there is no clear signal that tells the robot when the person it serves has changed their preferences~\citep{biyik2024preference}. The second gap is the coupling between different types of adaptation. A change in the body may require the perception module $\Phi$ to be updated because a new sensor changes what the robot can observe. It may also require the policy $\pi$ to be updated because the robot's control capabilities have changed. Yet the systems discussed above usually address one type of shift at a time, while a deployed humanoid may face several shifts at once, such as changes in payload, environment, and user preferences. This coupling means that self-evolving humanoids are not merely a special case of self-evolving software agents. They pose a broader embodied-systems problem.

\vspace{-0.1in}

\section{Self-Optimization in Humanoids}
\label{sec:Self-Optimization}

The third mechanism improves existing behaviors rather than learning new ones. As discussed in the previous section, self-optimization improves the policy $\pi$ using rewards and demonstrations through the world model and the planning workflow $\mathcal{W}$.
This section examines four components that support self-optimization. Reinforcement learning provides the improvement signal. The world model provides an internal simulator for evaluating candidate improvements. Planning breaks long tasks into smaller and achievable steps. Self-reflection identifies failures and suggests corrections. Section~\ref{sec:Self-Generateddata} then describes how these components form a data flywheel, where the experience collected by one policy becomes training data for the next.
A common theme connects these components. Every change made by a humanoid is eventually executed by its physical body. A candidate improvement must therefore be evaluated before it is deployed, and the world model provides the internal environment for this evaluation.

\begin{figure*}[t]
\centering
\includegraphics[width=0.9\textwidth]{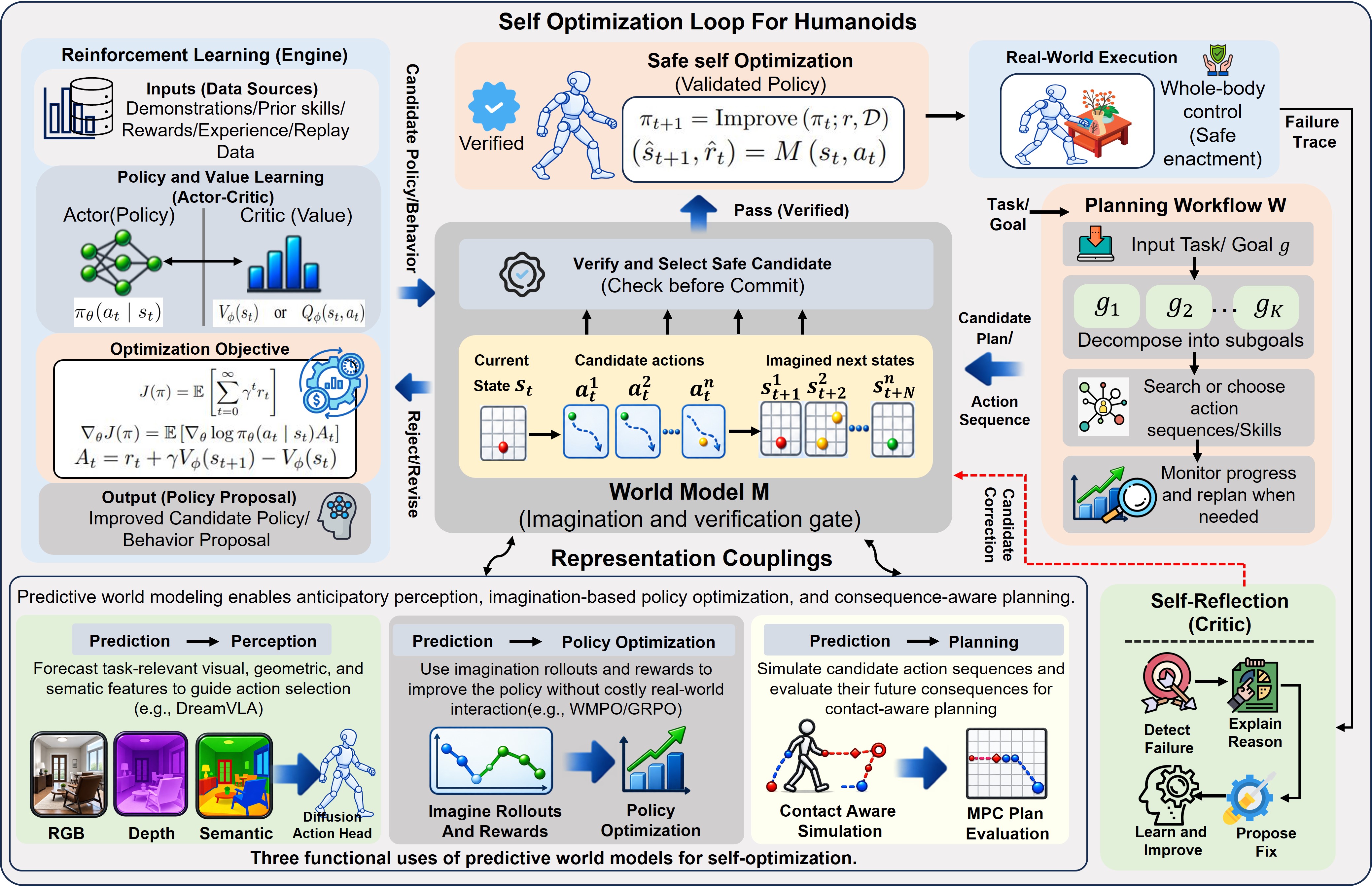}
\caption{
World-model-centered architecture for safe self-optimization in humanoid robots. RL, planning, and self-reflection generate candidate policy, action, and corrective updates, while the world model predicts their consequences before real-world execution~\citep{wu2022daydreamer,hansen2024tdmpc2}. Predictive world modeling couples to the self-optimization stack through three complementary pathways: prediction-to-perception~\citep{zhang2025dreamvla}, prediction-to-policy optimization~\citep{wmpo2025}, and prediction-to-planning~\citep{egovisionwm2025}, enabling anticipatory control, imagination-based learning, and consequence-aware planning.}
\vspace{-0.1in}
\label{fig:wm-hub}
\end{figure*}

\subsection{Reinforcement Learning in Humanoids}
\label{sec:Reinforcement}

Reinforcement learning is a main approach for improving the policy. Humanoid implementations build on standard policy-gradient and actor-critic methods~\citep{schulman2017ppo,haarnoja2018sac}, but differ in how much of the improvement process is learned.
HumanPlus~\citep{fu2024humanplus} implements this process from end to end. A low-level whole-body policy is first trained in simulation using 40 hours of retargeted human motion. The policy is then transferred to a 33-DoF, 180~cm humanoid, which can shadow a human operator in real time using a single RGB camera. The shadowing interface collects whole-body demonstrations, which are then used to train egocentric-vision skill policies. These policies complete tasks such as putting on a shoe, standing up, unloading a shelf, and folding clothes, with success rates of 60-100\% using at most 40 demonstrations.
OmniH2O~\citep{he2024omnih2o} extends this interface by using kinematic pose as a universal command space. A single student policy is trained by imitating a privileged teacher with large-scale motion retargeting. The resulting policy can be controlled through virtual-reality teleoperation, spoken instructions, an RGB camera, or a frontier model such as GPT-4o. Reference~\citep{cheng2024exbody} addresses the difficulty of applying imitation learning to high-dimensional humanoid whole-body control by encouraging the upper body to track a reference motion while relaxing constraints on the lower body.

ExBody2~\citep{exbody2_2024} focuses on improving the training procedure. It separates keypoint tracking from velocity control and uses a privileged teacher for distillation. This approach produces high-fidelity and expressive motion. The work also shows that policy quality depends on the balance between motion feasibility and diversity in the training dataset. The work of~\citep{mash2025} proposes a cooperative heterogeneous multi-agent approach to optimize locomotion of a single humanoid, which outperforms a single-agent approach in terms of system stability. SONIC~\citep{luo2025sonic} takes a different approach by focusing on scale. It trains a 42M-parameter tracking policy on more than 100 million motion-capture frames with 21,000 GPU hours of training. Robot, human, and hybrid motion commands are represented in a shared token space, allowing a single policy to support teleoperation, interactive navigation, and VLA-driven locomanipulation. BFM-Zero~\citep{bfmzero2025} is more closely related to self-evolution. It maps motions, goals, and rewards into a shared latent space, allowing a single policy to perform motion tracking, goal reaching, and reward optimization without retraining. The same latent space also allows a Unitree G1 to adapt from a small number of real-world examples and recover from disturbances without external intervention. For a self-evolving humanoid, the goal is not to find a good policy once, but to keep improving the policy as the body and environment change. BFM-Zero points toward such a design by allowing the same policy to adapt to different objectives and new experience without retraining.
More generally, a humanoid with behaviors organized in a structured latent space can improve after deployment by searching for better prompts rather than changing its policy weights. This makes adaptation inexpensive, easy to reverse, and easy to verify. The same principle is used in the preference-conditioned personalization discussed in Section~\ref{sec:personalization}. Over time, this form of improvement is closely related to reinforcement learning with verifiable rewards and self-play in software agents~\citep{yuan2024selfrewarding,zhao2025absolutezero,huang2025rzero}. However, a humanoid must improve while maintaining physical stability and staying within the limits of its body and environment.

\subsection{World Models in Humanoids}
\label{sec:World}

A world model is a learned predictor of dynamics. Given the current state of the system and a set of candidate actions, it predicts the outcome of each action. This capability makes self-optimization both more sample-efficient and safer, because the humanoid can evaluate the consequences of a behavior before executing it. Learning latent dynamics and training through imagined rollouts originates from early world-model methods~\citep{ha2018worldmodels}. DreamerV3~\citep{hafner2025dreamerv3} shows how this approach has matured. Specifically, a single configuration, stabilized through normalization, balancing, and transformation of learning signals, outperforms specialized algorithms on more than 150 tasks and learns to collect diamonds in Minecraft from scratch without human data. The key property is not this result itself, but the generality of the approach. An imagination engine that does not require per-domain tuning is important for an outer evolution loop, because no engineer will be available after deployment to retune the model as the environment changes. Two further results extend this approach in directions that are important for self-evolution. DayDreamer shows that a world model can also be learned directly on physical hardware~\citep{wu2022daydreamer}, while TD-MPC2~\citep{hansen2024tdmpc2} shows that the approach can scale. These works suggest that world models can benefit from scaling in a way similar to other foundation models. Genie~\citep{bruce2024genie} expands the type of data that a world model can use. It learns an 11B-parameter interactive environment from unlabeled internet videos and learns a latent action model without action labels. This suggests that a humanoid's world model does not need to rely only on data collected by the robot itself.

For humanoids, the world model and the policy can be coupled in three ways, as shown in Figure~\ref{fig:wm-hub}. DreamVLA~\citep{zhang2025dreamvla} couples prediction with \emph{perception}. Instead of predicting full frames, it predicts three types of information: dynamic regions, depth, and semantic content. These predictions provide compact look-ahead information for planning, and a diffusion action head uses this information to generate actions. This perception-prediction-action loop achieves 76.7\% success on a real robot. WMPO~\citep{wmpo2025} couples prediction with \emph{optimization}. It uses a pixel-space world model that is aligned with the web-pretrained visual features of the VLA policy. The world model serves as the environment for on-policy GRPO, allowing the policy to improve through RL without interacting with the real world. The authors report self-correction and continued improvement during extended training. This provides a direct example of policy improvement through imagination. The ego-vision contact world model of~\citep{egovisionwm2025} couples prediction with \emph{planning}. It is trained on an offline dataset without demonstrations and supports sampling-based model-predictive control with a learned value function. On a physical humanoid, the model enables real-time contact-seeking behaviors using only proprioception and egocentric depth. These behaviors include bracing against a wall after a shove, blocking an incoming object, and ducking under an arch. Open humanoid world-model approaches and dynamics-aware interaction models now target this setting directly~\citep{humanoidworldmodels2025,haic2026}. The 1X World Model Challenge has also started to standardize how these models are evaluated using real humanoid interaction data~\citep{1xworldmodel2025}. Within our framework, these systems serve the same role: they evaluate a candidate state $s_{\tau+1}$ before it is executed. This is the embodied counterpart of the empirical validation used by open-ended self-modifying agents~\citep{zhang2026dgm}. It is why we treat the world model as a component of safe self-evolution rather than only as a tool for improving efficiency. 

\begin{figure*}[t]
\centering
\includegraphics[width=1\textwidth]{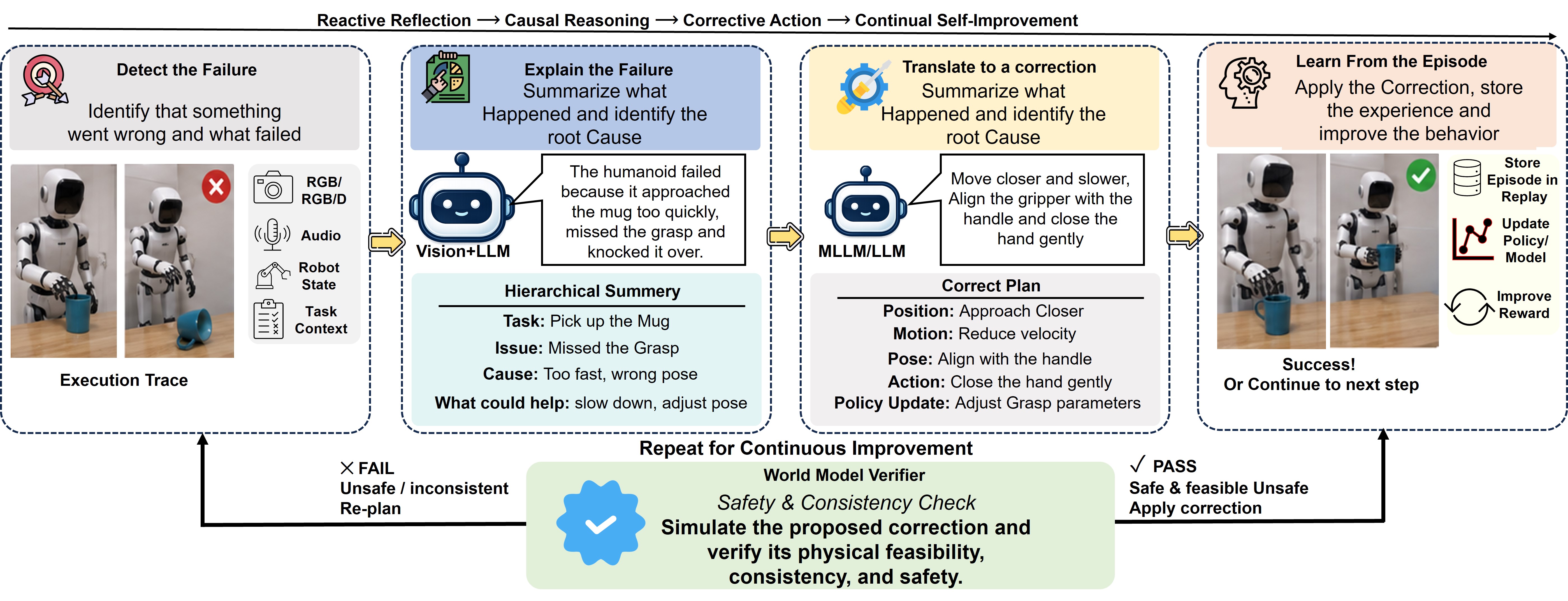}
\caption{
Self-reflection-driven self-optimization pipeline for humanoid robots. Failures are detected or anticipated during execution through progress- and sentinel-based monitoring~\citep{cyclevla2026,sentinelvla2025}, and then explained from multimodal execution traces~\citep{liu2023reflect}. The resulting explanation is translated into motor-level corrections~\citep{phoenix2025} or learning signals such as dense rewards~\citep{refadapt2025}, allowing the corrected experience to improve subsequent behavior through deployment-time learning~\citep{evolvevla2025,ttvla2026}. The world-model verifier evaluates its consistency, feasibility, and safety.}

\label{fig:reflection}
\end{figure*}

\subsection{Planning in Humanoids}
\label{sec:planning}

Planning is where the workflow $\mathcal{W}$ evolves: a long-horizon goal is first divided into sub-tasks, which are further decomposed when execution fails. SayCan~\citep{ahn2022saycan} introduced a language model that proposes possible next steps, and each step is weighted by the value function of the corresponding pretrained skill. The robot therefore selects actions based on both what it \emph{says} it should do and what it \emph{can} do from its current state. This approach enables a mobile manipulator to follow abstract long-horizon instructions. Humanoid systems extend this structure with the additional layers required for whole-body control. The hierarchical system in~\citep{hierarchicalvlp2025} places a vision-language planner above imitation-learned skills and an RL whole-body tracking controller. The planner selects skills and monitors their completion in real time. On 40 real trials of multi-step non-prehensile rearrangement with a Unitree G1, the system achieves 73\% full-sequence success. Proprioception-aware variants extend this structure to dual-arm settings~\citep{proprioplanning2025}, while classical motion and task-and-motion planning are surveyed in~\citep{rutili2024motionplanning}.

A more demanding question is whether these planners can handle failures during execution. DualTHOR~\citep{dualthor2025} studies this problem by extending AI2-THOR with dual-arm humanoid models, kinematic solvers, and physics-based failures during execution. The results show that current vision-language planners struggle with dual-arm coordination, and their performance decreases substantially when failures occur. The main challenge is therefore not only to generate a good plan at the beginning, but also to revise it when it no longer works. ThinkAct~\citep{thinkact2025} addresses this challenge by learning the planning process itself. A multimodal LLM is trained using visual rewards for task completion and trajectory consistency and generates reasoning-based plans. These plans are converted into a visual plan representation that guides a downstream action model. In simulation, the learned reasoning supports long-horizon planning and self-correction across a range of manipulation tasks. Together, these systems show how planning can move from a fixed program to a learned process that responds to feedback. The next step toward self-evolution is for the humanoid to improve this planning process using the failures it observes during its own operation. Related software-agent studies examine systems that improve their own workflows or designs ~\citep{zelikman2024stop,hu2025adas}.

\subsection{Self-Reflection in Humanoids}
\label{sec:self-reflection}
The most autonomous form of self-optimization allows the humanoid to evaluate its own behavior and make corrections. This idea originates from language agents, including verbal self-critique~\citep{shinn2023reflexion}, iterative self-refinement~\citep{madaan2023selfrefine}, and bootstrapped reasoning~\citep{zelikman2022star}. Embodied systems extend these ideas by grounding the critique in execution data. They can be organized into a four-stage pipeline as in Figure~\ref{fig:reflection}: \emph{detect} the failure, \emph{explain} its cause, \emph{translate} the explanation into a correction, and \emph{learn} from the episode. Early work focused mainly on the second stage. REFLECT~\citep{liu2023reflect} combines RGB-D, audio, and robot-state data to build a hierarchical summary of a failure. An LLM then uses this summary to identify the cause and generate a corrective plan. The RoboFail dataset released with the work provides data for studying this problem, and later datasets have expanded the available failure data~\citep{robofac2025}. However, explaining a failure is only part of the process. The robot must also convert the explanation into a physical correction. Phoenix~\citep{phoenix2025} addresses this problem using a multimodal LLM to convert the explanation into coarse motion instructions. A motion-conditioned diffusion policy then converts these instructions into precise actions. Finally, a dual-process motion adjustment mechanism supports efficient motion prediction and routes failures to a dedicated correction module. 

Failure detection is also shifting from reaction after the event to prediction before it. CycleVLA~\citep{cyclevla2026} adds subtask-progress estimation to a VLA policy and queries a VLM when the robot reaches a subtask boundary, where failures are more likely. When a failure is predicted, the system returns to an earlier step and retries using minimum-Bayes-risk decoding, which provides a form of test-time scaling for action models. Sentinel-VLA~\citep{sentinelvla2025} instead uses a lightweight sentinel to monitor execution continuously, reserving expensive reasoning for critical moments such as planning steps and detected errors. It also incorporates a continual-learning procedure that probes where the model performs poorly, collects data automatically in these difficult cases, and learns from them without human labeling. This procedure is analogous to human self-reflection, in which effort is concentrated on difficult problems while easier ones are set aside. The procedure gathered 2.6M transitions across 44 tasks and uses orthogonal adapters to limit forgetting, improving real-world success by more than 30\% over a strong VLA baseline.

The loop closes when the system uses its own reflection to improve the reward. Reflective Self-Adaptation~\citep{refadapt2025} converts a VLM explanation of a failure into a dense reward for on-robot RL. The method also keeps a success-based, quality-filtered imitation-learning pathway since the generated reward can be exploited by the policy. This highlights an important limitation of self-reflection: a system may generate a convincing explanation or reward that is still incorrect~\citep{yuan2024selfrewarding}. Test-time RL extends this idea by allowing the policy to improve during deployment. EVOLVE-VLA~\citep{evolvevla2025} replaces the oracle reward with a pretrained critic that measures task progress. It compares the current state with recent milestone frames and gradually increases the rollout horizon instead of comparing with the initial state. This achieves 8.6\% improvement on long-horizon tasks and 22\% improvement in one-shot adaptation. On unseen tasks, it reaches 20.8\% success, while supervised fine-tuning achieves zero success. TT-VLA similarly shows that dense progress-based rewards can improve a deployed VLA policy through test-time RL while preserving its pretrained knowledge~\citep{ttvla2026}. These results suggest that allowing a policy to continue learning during deployment can provide additional performance gains, similar to how additional inference-time compute can improve language-model performance. An advantage of this approach is that language-based corrections can be inspected. The reward-hacking result above indicates the corresponding risk: a confident but incorrect self-critique can degrade performance or compromise safety~\citep{macdiarmid2025natural}. Therefore, a world-model verifier should check the reflection loop before applying it.

\subsection{The Self-Generated Data Flywheel in Humanoids}
\label{sec:Self-Generateddata}

Self-optimization becomes a continuous loop when the experience collected during deployment is converted into training data for the next policy. The improved policy can then perform more capable tasks and collect better data, which becomes the training data for the following round. Recent work makes this loop increasingly explicit while reducing the human effort required at each step. The first step still depends on human operators, so the goal is to make data collection faster and less expensive. HOMIE~\citep{homie2025} combines an RL whole-body policy, controlled by a foot pedal, with an isomorphic arm exoskeleton and Hall-sensor gloves for hand control. Compared with earlier teleoperation systems, it reduces task completion time by about half and allows operators to reach more demanding configurations, including squatting and reaching high shelves. The authors present the system as a practical starting point for the self-improvement loop. Larger data-collection efforts, such as DROID and Open X-Embodiment, scale this approach by combining demonstrations collected across different environments and robot bodies~\citep{khazatsky2024droid,oxe2023}.

DART and DexHub~\citep{dexhub2024} move teleoperation to cloud-based simulation with augmented reality. This increases first-time operator throughput by 2.1$\times$ while reducing fatigue. Episodes can be uploaded to a shared public hub for reuse, and the resulting policies can still transfer to real robots. Data collection is therefore less dependent on access to physical robots, making the loop easier to scale. DexFlywheel~\citep{dexflywheel2025} automates the later stages of this loop, starting from a single human demonstration. Each cycle first uses imitation learning to learn the demonstrated behavior, then applies residual RL to improve and generalize it, and finally uses policy rollouts to collect new trajectories. These trajectories are augmented across different scenes and spatial configurations and used to train the next cycle. This produces more than 2{,}000 diverse demonstrations across four dexterous tasks. The resulting policies achieve 81.9\% success on held-out challenge settings and transfer to real hardware. The key property is that the policy trained in one cycle becomes the data generator for the next.

At the other end of the pipeline, GR00T~N1~\citep{groot2025} shows how diverse data can be combined to train capable robot models. Its dual-system VLA jointly trains a vision-language module and a diffusion-transformer action module using real-robot trajectories, human videos, and synthetic data. The resulting model outperforms imitation-learning baselines by an average of 11.6\% across different robot embodiments and has been deployed for bimanual humanoid manipulation with high data efficiency. In summary, these systems suggest that the self-optimization loop does not need to rely entirely on new human demonstrations. As the robot becomes better at generating useful experience, an increasing part of the data needed for further improvement can come from the robot itself. This leads to the fourth mechanism, self-generation.

\subsection{Section Summary and Takeaways}

Self-optimization relies on four components: RL provides the improvement signal~\citep{fu2024humanplus}, the world model acts as the verifier, planning provides the structure, and self-reflection serves as the critic~\citep{liu2023reflect}. For humanoid robots, however, none of these components should operate without a check. A change that improves performance in training can have physical consequences once it is applied to the hardware. Reward-driven and reflective updates should therefore be evaluated before they reach the physical system~\citep{wmpo2025}. We view the world model not only as a tool for faster learning but also as part of the safety mechanism.

A common pattern appears across these four components: useful improvements do not always require changing the model weights. BFM-Zero~\citep{bfmzero2025} searches for better prompts in a latent space, WMPO improves policies through imagined rollouts, while REFLECT and Phoenix~\citep{phoenix2025} express corrections in language before converting them into motor commands. These changes are relatively cheap to test, easy to reverse, and easier to inspect than direct weight updates. This architectural advantage is similar to the role of preference conditioning in personalization and also motivates the trust-region formulation discussed in Section~\ref{sec:safety}. However, this approach has two limitations. First, a verifier is only as reliable as the model behind it. A change that appears safe in simulation may still fail on real hardware. Second, a critic can be confidently wrong, as shown by the reward-hacking countermeasures discussed above~\citep{refadapt2025}. These limitations raise a more basic question: which changes should a humanoid be allowed to make? This question leads to self-generation, where the humanoid no longer relies solely on experience provided by the environment, but also creates the experience it needs.

\section{Self-Generation in Humanoids}
\label{sec:Self-Generation}
\begin{figure*}[t]
\centering
\includegraphics[width=0.95\textwidth]{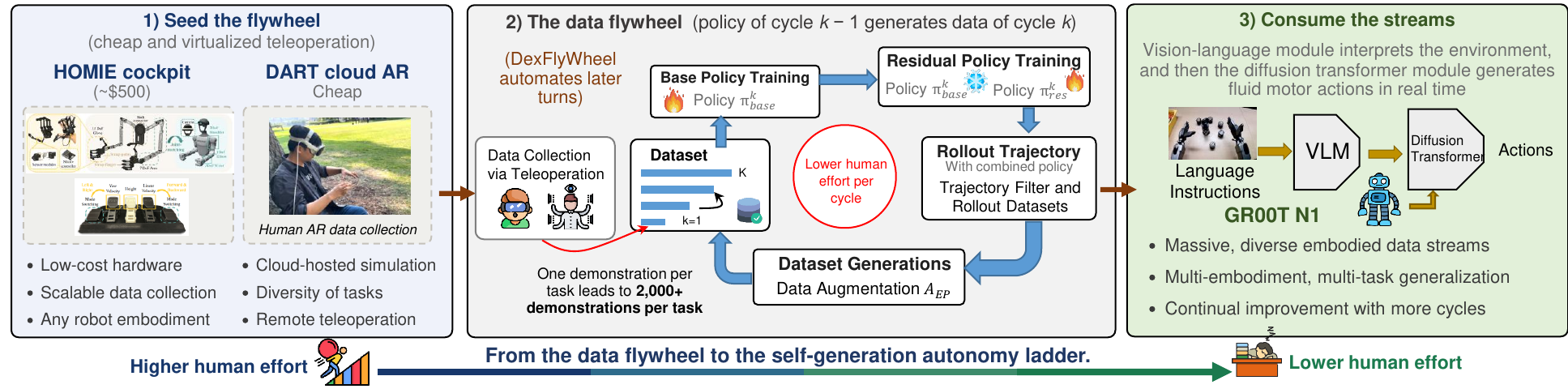}
\caption{From the data flywheel to the self-generation autonomy ladder. Moving from left to right requires progressively less human effort: HOMIE's cockpit~\citep{homie2025}, DART's cloud AR~\citep{dexhub2024}, DexFlywheel automated per-cycle data generation~\citep{dexflywheel2025}, and finally the GR00T~N1 foundation models~\citep{groot2025}.}
\label{fig:flywheel-ladder}

\end{figure*}

The first three mechanisms work with the experience provided by the environment: self-learning learns from it, self-adaptation responds to changes in it, and self-optimization uses it to improve behavior. Self-generation changes this role. Instead of waiting for useful experience to arrive, the humanoid creates the experience it needs. This includes demonstrations that were never collected, rollouts that are too slow or risky to run on hardware, and tasks and rewards that were not specified by an engineer. In our operator decomposition, self-generation corresponds to the operator $\mathcal{E}_{\mathcal{M}}$, which produces the stream $e_{\tau}$ used by the other three operators. The operator $\mathcal{E}_{\mathcal{W}}$ is also involved when the generated output is an objective or curriculum rather than data. We divide this literature into five approaches: synthetic data, diffusion-based generation, imagination, autonomous curriculum, and autonomous task discovery. We order these approaches by how much of the objective still needs to be provided by a human. Figure~\ref{fig:flywheel-ladder} summarizes these approaches together with the flywheel that connects them. This ordering also highlights the main trade-off in self-generation. Each step up the ladder gives the humanoid more control over the experience it obtains, while requiring less human input. At the same time, each step removes another human check from the loop. Self-generation therefore leads naturally to Section~\ref{sec:safety}, where we ask what changes a humanoid should be allowed to make to itself.

\subsection{Synthetic Data in Humanoids}
\label{sec:synthetic}

Human demonstrations are scarce in robot learning, and this family of methods seeks to use them more efficiently. The common procedure has three steps: an existing demonstration is transformed to fit a new situation, the transformed demonstration is replayed on the robot, and successful replays are kept as new demonstrations. Over time, the main change has been the reduction in human input required by this process. MimicGen~\citep{mandlekar2023mimicgen} established this pattern. It segments a source demonstration into object-centric parts, transforms each part to a new scene configuration, object instance, or robot arm, and then reconnects and replays the parts. With about 200 human demonstrations, MimicGen generated more than 50{,}000 demonstrations across 18 tasks. Policies trained on the generated data performed as well as policies trained on an equivalent number of \emph{additional human} demonstrations. This result suggests that synthetic amplification can reduce the need for collecting new demonstrations rather than simply supplementing them.

Later systems keep the same procedure but change the unit being transformed. DexMimicGen~\citep{jiang2025dexmimicgen} transforms whole trajectories for bimanual dexterous manipulation, where a human must teleoperate two arms and two multi-fingered hands at the same time. It groups tasks based on their coordination requirements, generates 21{,}000 demonstrations from 60 human demonstrations, and validates the generated data on a physical humanoid through a real-to-sim-to-real pipeline. SkillMimicGen~\citep{garrett2024skillmimicgen} instead uses the skill as the unit of transformation. It segments demonstrations into manipulation skills, adapts each skill separately, and reconnects them using motion-planned transit through free space. A hybrid skill policy learns when a skill starts, how it is executed, and when it ends, allowing a planner to re-sequence the skills at test time. The system generated 24{,}000 demonstrations from 60 human demonstrations. The resulting policies achieved an average success rate 24\% higher than the previous state-of-the-art~\citep{mandlekar2023mimicgen} and remained effective under clutter and large changes in the scene.

The remaining two systems focus on reducing the need for seed demonstrations. DemoGen~\citep{demogen2025} reduces the requirement to one human demonstration per task by editing both observations and actions. It adapts the demonstrated trajectory to new object configurations and synthesizes the corresponding \emph{observations} through 3D point-cloud editing. This approach supports deformable objects, dexterous hands, and bimanual settings. It can also produce behaviors that were not present in the original demonstration, such as resistance to disturbances. HumanoidGen~\citep{humanoidgen2025} removes the need for a human demonstration altogether. Given spatially annotated assets and a set of atomic dexterous operations, an LLM planner generates chains of spatial constraints that can be used to create bimanual demonstrations from scratch. Monte Carlo tree search is then used to plan over long horizons. On the accompanying benchmark, both 2D and 3D diffusion policies improved as more generated data was added. Other generators focus on specific interaction types, such as human-to-robot handover data~\citep{mobileh2r2025}. Open humanoid foundation models can then use these synthetic and video-generated data streams together with real trajectories during training~\citep{groot2025}. This is the final role of synthetic data: providing additional experience that can be used to train the next generation of humanoid policies.

In general, the approach shows a clear progression. The unit being transformed moves from the whole trajectory to the skill and finally to the individual observation, while the number of human seeds reduces gradually to zero. As a result, the seed-to-corpus amplification ratio increases from roughly $250\times$ to unbounded. This progression follows the trajectory predicted by the operator view: the operator $\mathcal{E}_{\mathcal{M}}$ produces $e_{\tau}$ from progressively fewer human seeds. It reflects the embodied form of self-play, where the system generates its own training problems~\citep{zhao2025absolutezero}. One point is important here: any guarantee that a generated demonstration is physically plausible comes from within the generator itself through its constraints, simulator, or a replay that either succeeds or fails, not from interaction with the physical world.

\subsection{Diffusion in Humanoids}
\label{sec:Diffusion}
Diffusion models can generate images, text, and video. Several studies extend this capability to humanoid robotics, either to generate experience data or to serve as control policies. In the self-generation setting considered here, diffusion creates new experience rather than directly controlling the robot. In self-generation, diffusion is used to create new experiences rather than directly control the robot. D-CODA~\citep{dcoda2025} illustrates this use. For eye-in-hand bimanual imitation, its diffusion model generates new wrist-camera observations that are consistent across viewpoints for both arms, together with the corresponding joint-space actions. A constrained optimization step then keeps the gripper-object contacts physically valid. Because the augmentation is performed directly in observation-action space, D-CODA can produce $e_{\tau}$ without building a full world model. It therefore sits between seed amplification and model-based imagination. A broader review of diffusion models in robotics is provided in~\citep{diffusionsurvey2025}. 

Diffusion models can also serve as the \emph{policy} itself. In this case, they model a multimodal distribution over actions rather than images. This use appears in Diffusion Policy~\citep{chi2023diffusionpolicy}, its 3D scene-conditioned extensions~\citep{ke20243ddiffuser}, and triply hierarchical visuomotor diffusion~\citep{lu2026h}. BeyondMimic~\citep{beyondmimic2025} shows how the same generative machinery can support both data generation and behavior generation. The method first learns a compact representation that can reproduce highly agile humanoid motions, including aerial cartwheels, spin-kicks, and sprinting, using a single set of hyperparameters. A unified latent diffusion model then represents these skills and uses \emph{classifier guidance}~\citep{dhariwal2021diffusion} at test time to steer the policy toward goals that were not specified during training. For example, it can complete a missing part of a motion or avoid an obstacle. These changes require no retraining, and the resulting behaviors transfer zero-shot to real hardware. Classifier guidance reflects a pattern that has appeared throughout personalization: the system changes its behavior at inference time, and these changes are cheap to test, easy to reverse, and can be constrained. Diffusion adds another useful capability: the same generative framework can produce both training data and robot behavior. This makes it a natural part of the generation stage in the self-evolution loop. However, this flexibility also has a limitation. A diffusion model can generate only what it has learned from its training distribution. Without additional constraints or interaction with the physical world, there is therefore no guarantee that a generated observation-action pair is physically valid~\citep{chi2023diffusionpolicy}.

\subsection{Imagination in Humanoids}
\label{sec:imagination}

Imagination generates new experience within a learned model instead of relying only on experience already collected by the robot. It supports self-evolution by allowing the humanoid to generate its own training experience. In this setting, the learned model acts as a substitute for the physical environment, allowing $\mathcal{E}_\mathcal{M}$ to run without additional physical interaction. Current systems mainly differ in how closely this imagined experience is tied to the scene in front of the robot. At the more tightly coupled end, DreMa~\citep{dreamtomanipulate2025} treats the world model as a \emph{learnable digital twin}. Gaussian splatting builds an explicit replica of the scene and represents its objects separately, while a physics simulator provides the dynamics. The robot can then imagine new object configurations by applying equivariant transformations, meaning that the demonstrated behavior moves consistently with the object. In their experiments, a Franka Emika Panda robot can learn new physical tasks using one example for each task variation.

Another approach includes models whose prior comes from internet-scale video. UniSim~\citep{yang2023unisim} shows that heterogeneous datasets can be combined into a single action-conditioned video simulator. Each dataset contributes the information it represents well: image data provides object variety, robot data provides actions, and navigation data provides motion. Both high-level vision-language policies and low-level control policies were trained entirely in the simulator and then deployed zero-shot in the real world. DreamGen~\citep{dreamgen2025} extends this idea to humanoids through a four-stage pipeline. First, an image-to-video model is adapted to the target robot. It is then prompted to generate photorealistic videos of familiar and novel tasks in diverse environments. Because these videos do not contain actions, pseudo-actions are recovered using either a latent-action model or an inverse-dynamics model. These pseudo-actions produce neural trajectories that can be used for policy training. A humanoid trained this way performed 22 \emph{new} behaviors in both seen and unseen environments, using teleoperation data from a single pick-and-place task in one environment.

DreamDojo~\citep{dreamdojo2026} applies this idea at the scale of a foundation model. It is pretrained on 44{,}000 hours of egocentric human video, making it one of the largest reported datasets for world-model pretraining. Because the videos do not contain action labels, the model represents actions as continuous latent variables that serve as a common representation across different action types. After further training on smaller robot datasets and distillation to 10.81 FPS, the model became fast and accurate enough for live teleoperation, policy evaluation, and model-based planning, including contact-rich tasks outside its training distribution. Between these two ends, the Unified Video Action model~\citep{unifiedvideoaction2025} combines the generator and policy in a single architecture. It uses masked training over a shared video-action latent space, allowing the same network to serve as a policy, video predictor, forward-dynamics model, or inverse-dynamics model depending on the task. Decoupled decoding also allows the model to predict actions without first generating the corresponding video, reducing inference latency.

Together with the compositional forecasting of DreamVLA~\citep{zhang2025dreamvla} and recent open humanoid world models~\citep{humanoidworldmodels2025}, these systems make imagination a productive way to generate new embodied experience $e_{\tau}$. They also make the operators $\mathcal{E}_{\mathcal{M}}$ and $\mathcal{E}_{\pi}$ in Eq.~(\ref{eq:operator}) more related to each other, since the same model can imagine possible experiences and evaluate candidate behaviors, as discussed in Section~\ref{sec:World}. The key technical step is learning pseudo-actions from unlabeled video. This allows video that the robot did not generate itself to enter the training loop as an approximation of robot experience. However, this also creates the main risk of model-based imagination. An imagined rollout is only as reliable as the model that generates it. A prediction error may not produce any real-world failure signal. As a result, a confident but incorrect update can reach the physical robot without being checked.

\subsection{Autonomous Curriculum in Humanoids}
\label{sec:Autonomouscurriculum}

The previous approaches generate experience using an objective that is fixed by a human. A curriculum extends this idea by organizing rewards and environments into a sequence that determines what the robot learns next. The sequence can be adjusted so that each new task remains achievable by the current policy. An autonomous curriculum moves this design into the learning loop: the humanoid generates the reward or environment that it will train against, evaluates the outcome, and revises it based on what it learns. In our decomposition, this corresponds to $\mathcal{E}_{\mathcal{W}}$ acting on the objective rather than $\mathcal{E}_{\mathcal{M}}$. Eureka~\citep{ma2024eureka} shows that reward design, long a step that depends heavily on expert knowledge in robot reinforcement learning, can be automated at a human-expert level. Given only the environment source code, an LLM generates candidate reward functions as executable code. An evolutionary loop then improves these rewards by training with each candidate, evaluating the results, and modifying the code based on the observed performance. Across 29 open-source environments and 10 robot morphologies, the generated rewards outperformed expert-designed rewards on 83\% of tasks, with an average normalized improvement of 52\%. When used as a curriculum, they also enabled the first simulated Shadow-Hand pen-spinning.

Two properties of Eureka matter beyond these results. First, its rewards are written as \emph{code}. Code can be read, compared line by line, and placed under version control, so a reward of this form is far easier to monitor and audit than an equivalent change buried in model weights. This is what makes Eureka relevant to the safety layer. Second, the evolutionary loop can take human feedback directly, without gradients. Eureka can therefore be viewed as a form of RLHF that improves the objective by editing reward code rather than by adjusting model parameters. Eurekaverse~\citep{eurekaverse2024} moves the same idea from the reward to the curriculum. Instead of designing a single reward, the LLM generates terrain programs that grow progressively harder while remaining diverse and solvable by the current policy. Environments and policies therefore evolve together: harder terrain pushes the policy to improve, and the improved policy makes still harder terrain tractable. The resulting parkour curriculum was transferred to a real robot and outperformed obstacle courses designed by humans. Taken together, the two systems show that an LLM can write either side of the training problem, the objective or the environment, as inspectable code.

A self-generated task is useful only if it is neither trivial nor unsolvable. Prioritized Level Replay~\citep{jiang2021plr} addresses this requirement by ranking generated levels according to their expected contribution to learning. Evolutionary Task Discovery~\citep{evotd2026} extends the criterion to skills, using crossover and mutation to explore combinations of skills and complexity, and filtering tasks by a zone-of-proximal-development criterion. The generated tasks therefore remain difficult enough to drive learning while staying within the current ability of the robot. These systems bring proposer-solver co-evolution~\citep{zhao2025absolutezero,huang2025rzero} to the physical robot.  The humanoid generates the rewards and environments that it must then solve, similar to how self-evolving reasoners generate their own problems. Embodiment provides one advantage and introduces one risk. The advantage is an audit trail: an objective written as code records what the robot was asked to optimize. The risk is that a reward that produces good performance may not reflect the intended objective. This is the same reward-hacking problem seen in reflective self-adaptation, but with an added difficulty: in this case, a human did not specify the objective in the first place.

\subsection{Autonomous Task Discovery in Humanoids}
\label{sec:Autonomoustask}
At the most autonomous end of the approach, the humanoid proposes its own tasks and skills. RoboGen~\citep{wang2023robogen} closes the loop with a self-guided propose-generate-learn cycle consisting of five stages. The agent first proposes tasks and skills to acquire. It then generates a matching simulation environment by retrieving and arranging suitable assets, decomposes each task into sub-tasks, and \emph{selects a learning method for each sub-task} from reinforcement learning, motion planning, or trajectory optimization. Finally, it generates the training supervision and learns the skills. Repeating this process allows the pipeline to produce a continuous stream of skill demonstrations with little human input. The fourth stage is particularly important because the system chooses the learning method itself rather than relying on an engineer to make that choice.

GenSim and GenSim2~\citep{wang2023gensim,hua2024gensim2} make task generation more structured by combining coding LLMs with multimodal reasoning. An LLM first converts an informal task description into an executable task definition that the robot can use for training. The systems generated up to 100 long-horizon tasks involving articulated objects and more than 200 objects in total. They also produced both planning and RL solvers that generalized across object categories. A proprioceptive point-cloud policy trained on these generated tasks transferred directly to real hardware without additional training. Co-training with a small amount of real-world data further improved performance by 20\% compared with training on real data alone. The self-evolving aspect is that the system generates both sides of the learning process: the LLM defines the task and selects the solver, allowing new capabilities to be learned without an engineer manually specifying either one. Beneath these language-driven systems is a more basic form of task generation. Unsupervised skill discovery learns diverse skills by maximizing the mutual information between skills and the states they visit, without requiring an external objective~\citep{eysenbach2019diayn}. Lifelong skill knowledge bases can then store these discovered skills so that they can be reused and composed later~\citep{skillscrafter2026}. Together, these methods represent the embodied form of agents that propose their own tasks and skills~\citep{yuan2025evoagent,wang2024voyager}. The key difference is the cost of an open-ended proposal when it is executed on a physical robot. Since the objective is not specified by a human, this approach makes the verification question in Section~\ref{sec:safety-constraint} particularly important.

\subsection{Section Summary and Takeaways}

Self-generation completes the loop started by the other three mechanisms. The synthetic data and imagined rollouts it produces become new experience for self-learning, while its generated rewards and curricula provide the objectives that self-optimization can improve. We therefore view the four mechanisms as stages of a single cycle. Across the five approaches above, the role of the human becomes progressively smaller. The progression moves from learning with hundreds of demonstrations~\citep{jiang2025dexmimicgen}, learning from a single demonstration~\citep{dreamtomanipulate2025}, learning without demonstrations~\citep{humanoidgen2025}, to generating the reward, and finally to proposing tasks that no human has defined. As this happens, the responsibility for checking what the robot generates also shifts toward the system itself. Instead of relying on human judgment, the robot uses its own reward functions, sampling constraints, simulated physics, or world-model predictions to decide whether the generated experience is useful. 

This shift from human to robot judgment is easier to manage when the generated artifact is transparent, such as a reward function written in code~\citep{eurekaverse2024}. It is harder to manage when the artifact is difficult to interpret, such as a diffusion-generated sample~\citep{chi2023diffusionpolicy}. The main concern is not whether the generated data are statistically accurate, but whether they are physically safe. A curriculum that rewards unsafe behavior, or a world model that predicts contact incorrectly~\citep{egovisionwm2025}, can become a hazard once the resulting behavior reaches the physical robot. Self-generation therefore raises a further question, beyond what a humanoid can change about itself: what changes should it be allowed to make?

\begin{figure*}[t]
\centering
\includegraphics[width=0.95\textwidth]{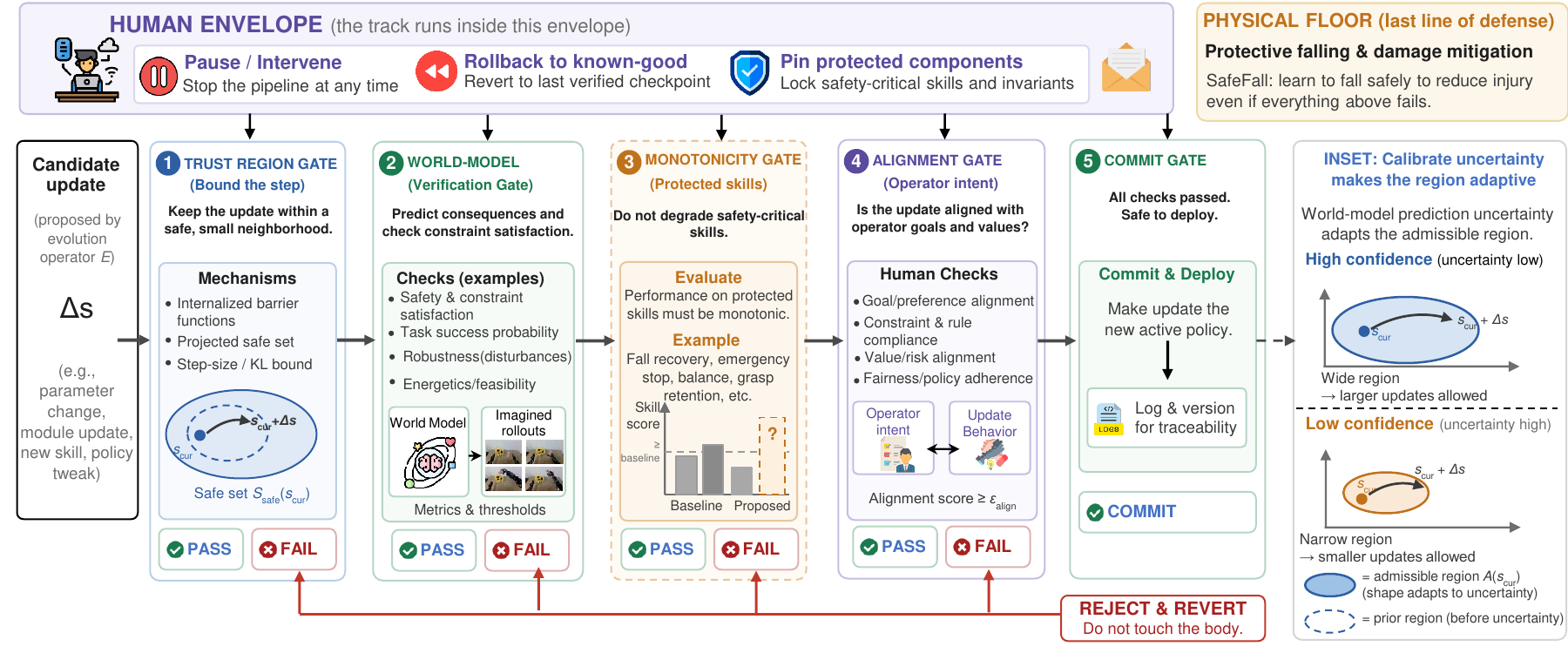}
\caption{A candidate update $\Delta s$ proposed by the evolution operator must pass five gates in sequence: a trust region bounding the step, the world-model verification gate~\citep{dreamgen2025}, a monotonicity check on protected safety-critical skills, an alignment check against operator intent, and only then commit. The whole track runs inside the human envelope, with protective falling SafeFall~\citep{safefall2025} as the physical floor beneath it.}
\label{fig:gauntlet}
\end{figure*}

\section{Safe and Uncertainty-Aware Self-Evolution}
\label{sec:safety}
The four mechanisms discussed so far focus on capability: what a self-evolving humanoid can learn, change, and improve. This section asks a different question: what should the robot be allowed to change, and what evidence should it provide before making that change? Embodiment makes this question more important. For a physical robot, an unsafe update is not simply a loss of performance; it can become a hazard to people nearby. At the same time, a humanoid that can modify its policy, perception, memory, workflow, and body model cannot have every possible change specified in advance. The same openness that makes self-evolution powerful also makes it difficult to constrain. This risk is not merely theoretical. The first systematic study of misevolution shows that self-evolving agents can accumulate risk through four pathways: the model, memory, tools, and workflows~\citep{shao2026misevolution}. The study found that misevolution can occur even in agents built on strong foundation models. Safety alignment can gradually degrade as self-generated memories accumulate, while agents can also introduce vulnerabilities unintentionally when they create and reuse their own tools.

In our framework, the model pathway corresponds to updates to the parametric components $\pi$ and $\Phi$, while the memory pathway corresponds to $\mathcal{E}_{\mathcal{M}}$. The tool and workflow pathways correspond to $\mathcal{E}_{\mathcal{W}}$ and the skill interfaces stored in $\mathcal{M}$, as in Eq.~(\ref{eq:state}). However, none of these pathways cover the body operator $\mathcal{E}_{\mathcal{B}}$. This omission is important for humanoids. It comes from the setting in which the taxonomy was developed: a software agent has no physical body that can drift. A stale or corrupted body model can give the robot an incorrect view of its own dynamics. This error can then affect every other safety mechanism, because a verification procedure is only as reliable as the body model used in the simulation. Body drift is therefore a fifth misevolution pathway that is specific to embodied systems. We therefore require the evolution operator $\mathcal{E}$ to remain within an \emph{admissible region} $\mathcal{C}(s_{\tau})$. This region contains the states and updates that satisfy the robot's safety, verification, monotonicity, and alignment requirements. The constraint $s_{\tau}\in\mathcal{S}_{\mathrm{safe}}$ in Eq.~\eqref{eq:lifelong} is therefore no longer an abstract condition. It becomes an explicit requirement on what the humanoid is allowed to change during self-evolution.

\subsection{Safety as a Constraint on Admissible Evolution}
\label{sec:safety-constraint}

We organize the admissibility mechanisms in the literature into five layers of defense, shown as a single update passing through the gauntlet in Figure~\ref{fig:gauntlet}. We synthesize this five-layer organization. Each layer draws on an existing line of work, except monotonicity, which we introduce as a safety interpretation of catastrophic forgetting. The layers are ordered from the cheapest and most general test to more expensive checks. This allows an update that moves the system too far to be rejected before it reaches the simulator.

A \emph{trust region} limits how far a single update can move the robot's state. If an update turns out to be harmful, the change should be small enough to detect and reverse. This is directly related to the plasticity cost $\rho_{\mathrm{plas}}$: an update should be penalized when it changes the system faster than the verification mechanisms can determine its effects.
Humanoid work applies this principle at three levels. CBF-RL~\citep{cbfrl2025} embeds the safety constraint in training itself, where a control-barrier-function term restricts deviation from the nominal RL policy and safety-filtered rollouts confine exploration to safe limits. A Unitree G1 learned to climb stairs and avoid obstacles without a separate runtime safety filter. The property matters for self-evolution, in which exploration may become progressively more aggressive.
When many constraints interact or conflict, as in whole-body collision avoidance inside a cabinet, the projected safe-set method~\citep{dexteroussafe2025} relaxes the constraints while minimizing violations, which keeps a feasible safe control available. At a higher level, an interpretable template-based foothold planner can be combined with an RL tracker and a learned foothold modifier. A G1 crossed a 0.2~m-wide beam twenty times without failure under this arrangement~\citep{narrowpaths2025}. The principle in both cases is to keep an interpretable physics-based component outside the learned policy, providing a stable reference that the later layers can protect.

A \emph{verification gate} checks a candidate state $s_{\tau+1}$ before it reaches the physical robot, either in a physics simulator or a learned world model. This is why we treat world models as part of the safety machinery of self-evolution rather than simply as tools for faster learning. However, the verification is only as reliable as the model used for the check. As the physical robot drifts away from the model, the reliability of the verification can decrease. Catastrophic forgetting is therefore both a learning and a safety issue. An update may improve overall performance while removing a rarely used behavior that is still critical for safety.

We therefore introduce a \emph{monotonicity} constraint that prevents regression on a protected set of safety-critical metrics. This constraint addresses an important tension in lifelong adaptation. Some forgetting is necessary because outdated body or world knowledge must be replaced, but knowledge that is important for safety should not be lost. To our knowledge, no existing humanoid system explicitly enforces this type of monotonicity. We therefore view it as a necessary safety layer that remains largely an open problem rather than a solved one.

An \emph{alignment} check keeps the humanoid's evolving objectives tied to operator and user intent. It is the embodied form of aligning a model with human feedback~\citep{christiano2017deeprlhf,ouyang2022instructgpt}, which also supports the personalization discussed in Section~\ref{sec:personalization}. This check is especially important for the self-authored rewards and curricula in Section~\ref{sec:Self-Generation}. These objectives are generated by the robot rather than written by a human, so they may never have been directly examined by one.

A \emph{human envelope} keeps the ability to pause evolution, roll back to a known-good state, and protect critical components, such as a validated locomotion controller or a hard safety rule, from modification. This layer differs from the other four. It is not a gate that a candidate update must pass. Instead, it defines the boundaries within which the entire evolution process operates. It is also the only layer that remains available after an update has been committed.

ShieldAgent~\citep{chen2025shieldagent} suggests what such a mechanized envelope could look like. It is a guardrail agent that enforces explicit safety policies on the trajectory of another agent. It compiles policy documents into verifiable, action-conditioned rule circuits and formally checks trajectories against them. It improved over prior guardrails by 11.3\%, reaching 90.1\% recall. A humanoid version could compile robot-safety standards into machine-checkable constraints on proposed \emph{updates}, rather than only on actions. This would complement the classical safe-RL toolbox~\citep{garcia2015saferl}. The remaining gap is clear: existing standards are designed for machines with fixed and certified behavior, not for machines that can evolve themselves.

Beneath the five layers is a physical line of defense for cases where all the checks above have failed. SafeFall~\citep{safefall2025} runs a lightweight recurrent fall predictor alongside the main controller and activates a protective RL policy only when a fall is judged imminent and unavoidable. The protective policy increases the time before impact, distributes contact forces, and protects vulnerable parts of the robot. It was tested on a full-scale humanoid during rope-induced falls at 3~m/s. Curriculum-based methods can also discover protective postures suited to the robot's morphology without relying on hand-designed human priors~\citep{selfprotectivefall2025}. We treat this as a safety floor, not a sixth admissibility layer, because it does not decide whether an update is allowed. Instead, it limits physical damage when an accepted update later proves wrong. This is especially important for self-evolving humanoids, which can make updates that no engineer has reviewed and whose verification depends on a world model that may become inaccurate as the robot changes.

\subsection{Uncertainty-Aware Models}
\label{sec:safety-models}
Safe evolution requires the humanoid to estimate and calibrate its uncertainty. It can then adapt more aggressively when confidence is high and more cautiously when confidence is low. Gaussian processes~\citep{rasmussen2006gp} provide a calibrated posterior variance that can be used directly in the safety check. The Gaussian Control Barrier Functions (CBF) in~\citep{khan2022gpcbf} show how this can be done. Instead of requiring a barrier function to be specified in advance, it learns the CBF \emph{online} from safety samples under a GP prior. The quantities needed for safe control can then be obtained analytically. The resulting safe set can be non-convex and can be \emph{updated} as new data arrives. This ability to update the safe set is important for self-evolving humanoids. Both the robot body and its environment can change over time, so the safe set should be learned and updated rather than fixed once. Uncertainty-aware predictive CBFs extend this idea to human-robot interaction~\citep{uapcbf2025}. They combine predictions of human hand motion with formal safety guarantees, tightening the safety margin when the prediction is uncertain and relaxing it when confidence is higher. This avoids the overly conservative worst-case envelopes that can limit human-robot interaction.

Uncertainty can also guide what the humanoid asks rather than what it does. Since each pairwise comparison provides at most one bit of information, active preference-based reward learning~\citep{biyik2024preference} models the reward over trajectory space and selects the comparison that provides the most information. This makes preference-based personalization more effective with only a small number of queries. Bayesian deep learning and deep ensembles~\citep{lakshminarayanan2017ensembles,gal2016dropout} extend uncertainty estimation to the high-dimensional perception and world models used by humanoids, although exact calibration remains difficult. Model-predictive out-of-distribution detection can also identify when the live input has moved outside the training distribution, at rates suitable for humanoid operation~\citep{rapt2026}. In our framework, uncertainty makes the admissible region $\mathcal{C}(s_{\tau})$ \emph{adaptive}: the trust region contracts when the world model is uncertain about a candidate update and expands as confidence increases. For example, consider a humanoid whose gripper has been replaced. Since its world model was trained with the previous hand, grasp-related updates will have high uncertainty. The trust-region gate should therefore allow only a small correction, which is re-verified after a few episodes. As the robot collects more experience with the new gripper and predictive uncertainty decreases, the same gate can allow larger updates. This is the embodied counterpart of agents that update only on uncertain or informative samples~\citep{acikgoz2025ttsi}. The difference is that, for a humanoid, a miscalibrated uncertainty estimate can determine how far the physical robot is allowed to move before the next safety check.

\begin{table*}[t]
\centering
\caption{Representative benchmarks, simulators, and datasets for humanoid learning.}
\label{tab:benchmarks}
\small
\setlength{\tabcolsep}{4pt}
\scalebox{0.8}{
\begin{tabular}{llll}
\toprule
\textbf{Resource} & \textbf{Type} & \textbf{Sim / real} & \textbf{Targets lifelong/drift} \\
\midrule
HumanoidBench~\citep{sferrazza2024humanoidbench}      & Whole-body benchmark & Sim  & No \\
LIBERO~\citep{liu2023libero}                          & Manipulation benchmark & Sim & Lifelong task suites \\
MimickingBench~\citep{mimickingbench2024}             & Scene-interaction benchmark & Sim & No \\
Humanoid Everyday~\citep{humanoideveryday2025}        & Manipulation dataset & Real & No \\
Genie Sim~\citep{geniesim2026}                        & Simulation platform & Sim & No \\
Isaac Gym / Isaac Lab~\citep{makoviychuk2021isaacgym,mittal2025isaaclab} & Simulator & Sim & No \\
SIMPLE~\citep{simple2026}                             & Loco-manip.\ sim + eval & Sim & No \\
Open X-Embodiment / DROID~\citep{oxe2023,khazatsky2024droid} & Cross-embodiment dataset & Real & No \\
AMASS / PHUMA~\citep{mahmood2019amass,phuma2025}      & Motion dataset & Mocap / real & No \\
\bottomrule
\end{tabular}}
\end{table*}

\section{Evaluation, Metrics, Benchmarks, and Datasets}
\label{sec:evaluation}

The survey so far has described what a self-evolving humanoid can do and what it should be allowed to do. This section turns to how such a system should be measured. Without benchmarks that measure how a humanoid changes over time, it is difficult to determine whether a theoretical guarantee holds, whether a system can sustain self-evolution in practice, or whether an evolved behavior remains safe. We therefore view the lack of a benchmark specifically designed for self-evolving humanoids as the most important gap in this survey.

\subsection{From a Checkpoint to a Trajectory}
\label{sec:eval-principles}

Standard benchmarks are not well suited to systems that can change themselves and, in doing so, change the data they learn from. A fixed test set measures how the policy $\pi_{\tau}$ performs at one point in time, but says little about how the robot adapts when its environment, tasks, or internal state change. The problem is even harder for a self-generating humanoid because the robot can partly decide which experiences to create and which tasks to attempt. A fixed benchmark may therefore give a misleadingly high score if the robot avoids situations in which it is likely to fail. Evaluation should instead focus on the \emph{evolution process itself:} how the system changes over time, how well it adapts to different types of drift, and whether these changes improve or degrade its lifelong performance $V$.

The agent literature has started to make this shift. The experience-driven lifelong-learning framework of~\citep{cai2025lifelongbench} evaluates agents over simulated lifetimes with interdependent tasks that span months, rather than over isolated episodes. The misevolution study further argues that safety should be measured over time because it can degrade as the agent performs self-driven updates~\citep{shao2026misevolution}. Surveys of agent evaluation also emphasize process-level rather than only outcome-level measurement~\citep{yehudai2025agenteval}. Humanoid systems inherit these requirements and add the physical body. Drift scenarios should therefore include changes in the robot's morphology, while the protected metrics should also cover physical-safety behaviors. The metrics defined below address both aspects.

\subsection{Metrics for Self-Evolving Humanoids}

\label{sec:eval-metrics}
Evaluation over the whole trajectory requires metrics that capture what happens along it. We therefore define four metrics that can serve as a reporting standard for the benchmark proposed below. Let $\bar{u}_{\tau}$ denote the average task utility during the fast loop at slow step  $\tau$, let a drift event occur at $\tau{0}$, and let $a_k(\tau)$ denote the competence of the robot on a previously mastered skill $k\in\mathcal{K}$. We propose reporting the following four quantities.
\begin{equation}
\begin{aligned}
\eta(\tau_0) &= \Delta^{-1}, \qquad
\Delta = \min\left\{\delta:
\bar{u}_{\tau_0+\delta}
\geq (1-\epsilon)\bar{u}_{\tau_0^-}\right\},\\
\varphi(\tau) &= \frac{1}{|\mathcal{K}|}
\sum_{k\in\mathcal{K}}
\left[
\max_{\tau'\leq\tau} a_k(\tau')-a_k(\tau)
\right],\\
A(\tau_0) &= \sum_{\delta=0}^{\Delta}
\left(
\bar{u}_{\tau_0^-}-\bar{u}_{\tau_0+\delta}
\right),\\
\sigma(T) &= \frac{1}{T}
\sum_{\tau\leq T}
\mathbb{1}\!\left[
s_\tau\notin\mathcal{C}(s_\tau)
\right].
\end{aligned}
\label{eq:metrics}
\end{equation}
The \emph{adaptation rate} $\eta(\tau_{0})$ measures how quickly the robot recovers after a change, as the inverse of the time required to return to $(1-\epsilon)$ of its pre-drift utility. The \emph{forgetting rate} $\varphi(\tau)$ measures how much previously acquired capability has been lost, and is the embodied counterpart of backward transfer in continual learning. The \emph{transient cost} $A(\tau_{0})$ is the utility forgone while adapting, which a snapshot evaluation does not capture. The \emph{safety-regression rate} $\sigma(T)$ measures how often the robot violates its protected constraints over the evaluation period. This last quantity has no direct analogue in purely software-based learning systems, because the consequences of a violation are physical.
These metrics also permit the trade-off between adaptation and stability to be studied directly. Varying the  $\rho_{\mathrm{plas}}/\rho_{\mathrm{stab}}$ balance and plotting the resulting $(\eta,\varphi)$ pairs yields the \emph{plasticity-stability frontier} of a system. All four metrics are either direct components of the lifelong objective $V$ or constraints on it, and therefore provide a computable surrogate for a quantity that is otherwise difficult to measure. They build on existing work on forgetting and knowledge transfer in continual and lifelong robot learning~\citep{meng2025legion,liu2023libero}, and extend the drift-recovery perspective demonstrated on physical robots in~\citep{wu2022daydreamer}.

Two lessons follow from this line of work. First, evaluating a system over its full trajectory can reveal behavior that a single snapshot misses. LIBERO showed that plain sequential fine-tuning can outperform purpose-built lifelong-learning methods in forward transfer while performing worse in retention. It also showed that naive supervised pretraining can hurt subsequent lifelong learning~\citep{liu2023libero}. Metrics of this kind keep these trade-offs visible instead of reducing them to a single score. Second, trajectory-based metrics require \emph{attribution}. The torque variation score of~\citep{imitationdifficulty2025} separates imitation errors caused by a weak policy from those caused by the intrinsic difficulty of the target motion. The metrics $\eta$ and $\varphi$ need a similar decomposition. For example, slow recovery may result from either a weak evolution operator or a particularly difficult drift event. Only the former should be attributed to the system.

\subsection{Benchmarks, Simulators, and the Missing Benchmark}

\label{sec:eval-benchmarks}

Table~\ref{tab:benchmarks} summarizes the infrastructure available for studying self-evolving humanoids. Three entries are especially relevant to the argument. HumanoidBench covers 27 whole-body locomotion and manipulation tasks on a Unitree H1 with two dexterous Shadow Hands. Its main finding is diagnostic: standard RL performs poorly on most tasks, while hierarchical agents built on robust low-level skills perform better~\citep{sferrazza2024humanoidbench}. This result highlights the capability gap that self-optimization must address. Humanoid Everyday provides 10.3K real teleoperated trajectories across 260 tasks, with RGB, depth, LiDAR, and tactile data. It also provides a cloud platform for deploying and evaluating external policies under controlled conditions~\citep{humanoideveryday2025}. PHUMA shows that \emph{data quality can also be measured}: its physics-aware curation detects and removes joint-violation, floating, penetration, and skating artifacts. Policies trained on the resulting 73-hour dataset tracked motions more accurately than baselines trained on AMASS and Humanoid-X~\citep{mahmood2019amass,phuma2025}. The remaining entries provide benchmarks for humanoid-scene interaction~\citep{mimickingbench2024}, infrastructure for transfer-oriented training~\citep{humanoidgym2024}, cross-embodiment datasets~\citep{oxe2023,khazatsky2024droid}, and GPU-parallel simulators and manipulation suites that provide the required training throughput~\citep{makoviychuk2021isaacgym,mittal2025isaaclab,robocasa2024,maniskill3_2024,maniskillhab2024,simple2026}.

Two recent platforms approach self-evolution from different directions. GaussGym integrates Gaussian-splatting rendering with vectorized physics at over 100{,}000 steps per second and accepts iPhone scans, scene datasets, and generative-video outputs~\citep{gaussgym2025}. This allows a photorealistic replica of the \emph{actual deployment site} to serve as a training and evaluation environment. TTT-Parkour uses this real-to-sim capability for adaptation. Genie Sim 3.0 takes a different approach by automating evaluation. It combines an LLM-based scene generator with a VLM-based assessment pipeline to create more than 100{,}000 evaluation scenarios and over 10{,}000 hours of synthetic data~\citep{geniesim2026}. This makes it possible to evaluate systems continuously rather than relying only on hand-graded checkpoints. However, neither platform directly evaluates \emph{self-evolution itself}. LIBERO comes closer by treating lifelong robot learning as knowledge transfer across a sequence of tasks and allowing a generator to produce an open-ended stream of new tasks~\citep{liu2023libero}. The experience-driven lifelong benchmark makes a similar shift for software agents by evaluating the learning process over time~\citep{cai2025lifelongbench}. What is still missing is a benchmark that asks the central question for humanoids: can the robot improve from its own post-deployment experience while adapting to changes in both its environment and its own body?

Such a benchmark does not need to be built from scratch because most of the required components already exist. GaussGym provides realistic environments and real-to-sim reconstruction, Genie Sim 3.0 provides automated trajectory-level evaluation, PHUMA provides metrics for filtering and measuring unsafe or low-quality motion data, and Humanoid Everyday provides a platform for standardized policy evaluation. Combining these components with controlled drift scenarios and a protected set of safety-critical behaviors is therefore mainly an engineering task. We see this as one of the most valuable next steps for the humanoid domain. Without such a benchmark, it remains difficult to compare self-evolving humanoids or determine the efficiency of the discussed mechanisms.

\section{Open Challenges and Future Directions}
\label{sec:challenges}
 
Having established how self-evolution should be measured, we now turn to what the field cannot yet do. We group the remaining open problems into three areas: theory, systems, and governance. For each area, we identify the gap revealed by the preceding sections, explain why it is difficult, and suggest a concrete direction for future work. All three areas depend on the trajectory-level measurement discussed in Section~\ref{sec:evaluation}, because when a property cannot be proved, it must instead be monitored throughout the robot's lifetime.

\subsection{Foundational and Theoretical Challenges}
\label{sec:ch-theory}

The deepest open question is whether post-deployment self-improvement can be made \emph{provably} monotone on a physical robot. The idea of a machine that changes itself only when the change can be proved beneficial~\citep{schmidhuber2006godel} is not practical, and recent self-improving systems instead rely on \emph{empirical} validation~\citep{zhang2026dgm,wang2026hgm}. This makes the metrics more than descriptive: when a property cannot be proved, it must be measured repeatedly throughout the robot's lifetime. Three concrete problems follow. The first is \emph{sample complexity}: how much experience does a humanoid need to keep up with changes in both its environment and its own body? If adaptation requires too much data, lifelong self-improvement may not be practical in real deployments. The second is finding a principled and \emph{learnable} way to forget outdated body and world knowledge without losing skills that remain useful. This is one of the clearest open problems and connects directly to the monotonicity layer of our safety framework. A good forgetting mechanism could serve both as a learning rule and as evidence that safety-critical knowledge is preserved. The third is \emph{convergence}. The trade-off between plasticity and stability can be formulated as an optimization problem, but we do not yet know when a system will settle into a useful balance rather than oscillate between too much and too little adaptation. A trajectory-level $(\eta,\varphi)$ frontier provides a way to reveal this behavior.

\subsection{Systems and Embodiment Challenges}
\label{sec:ch-systems}

For self-evolution to be useful in practice, it must eventually run on the robot itself. The full propose-simulate-validate-remember loop is computationally expensive, and running it for years within real-time, energy, thermal, and memory limits remains an open problem. A two-timescale design provides part of the solution because the slower evolution loop does not need to run at every control step. The harder problem is how to schedule this limited computation while preserving enough resources for normal operation. Two factors make this scheduling problem difficult. The first is the sim-to-real gap. A world model or simulator may be sufficient to test a candidate update in imagination, but differences between simulation and the physical robot can still cause the update to fail on hardware~\citep{worldmodelsurvey2025,ding2025worldmodelsurvey}. This limits how much of the validation process can be moved off the robot. The second is that the robot itself changes. Hardware wear alters its dynamics and becomes a source of drift that the self-model must detect and track rather than treat as external noise~\citep{hu2025simself}. As the body changes, the model used to verify new behaviors becomes less accurate, bringing the scheduling problem back to the body operator. Orchestration adds another challenge. A humanoid does not rely on a single learned model. It may maintain separate models for perception, control, planning, memory, and world prediction at the same time. Deciding which components to update, when to update them, and in what order is difficult because a change in one component can alter the conditions under which the others operate. A poorly scheduled update may therefore destabilize the evolution loop even when each individual update is admissible. This problem is especially important in real deployments, where changes rarely occur one at a time. A robot may encounter a new environment and a changed payload during the same period. Handling such coupled changes within the robot's physical and computational limits remains a central requirement for self-evolving humanoids.

\subsection{Safety, Governance, and Human Oversight}

\label{sec:ch-governance}

Certifying a self-evolving humanoid means certifying a \emph{process} that can produce new behaviors rather than a fixed behavior. Existing assurance practices are not designed for this setting. Runtime shielding and verifiable safety policies address part of the problem during execution~\citep{chen2025shieldagent}, but here the unit of approval is the update rather than the individual action. Three gaps follow. First, robot-safety standards are designed for machines whose behavior is fixed and certified before deployment. They provide no clear way to handle a machine whose behavior changes after deployment or to recertify it as it evolves. Second, there is no agreed way to audit an evolved change after the fact. Such an audit requires a record of what changed and how the change affected the system. This connects measurement with mechanism design. As seen throughout the survey, changes are easier to audit when they occur outside the model weights, such as through code-level rewards, adapters, or latent prompts. It also supports continuously logging $\sigma(T)$ and the protected set rather than checking them only at selected points. Third, responsibility for physical harm caused by an autonomously evolved behavior remains unclear. It is not yet clear whether this responsibility should fall on the manufacturer, the operator, or the evolution mechanism itself. Resolving this issue requires the first two gaps to be addressed: there must be both a record of what changed and a standard against which the change can be evaluated. The misevolution study~\citep{shao2026misevolution} shows that this is a practical concern

\section{Conclusion}
\label{sec:conclusion}
Humanoid robots are built to operate for years, yet most current systems stop learning once they leave the laboratory. This limitation becomes more important as deployment expands. Environments change, tasks evolve, users have different needs, and the robots themselves degrade through wear, damage, and aging. A policy that works well at deployment can therefore become less effective over time. In this survey, we model a deployed humanoid as a state that includes its policy, perception, memory, workflow, and physical body. Post-deployment improvement is then treated as a process that continually updates parts of this state toward a lifelong objective.  The physical dimension makes lifelong self-evolution in humanoids a more challenging problem than in software agents. Across the mechanisms surveyed, recent systems place post-deployment changes in prompts, adapters, external memory, and reward functions, and these components are relatively easy to test and reverse. As human supervision becomes less frequent, systems increasingly rely on world models, simulators, automated evaluation, and self-generated experience to decide whether an update is useful. Therefore, verification also becomes a central problem in self-evolution. We consequently treat safety not as a property of a single policy, but as a property of the evolution process itself. We organize existing safety mechanisms into five layers, from trust regions that limit individual updates to a human-defined envelope that bounds the entire process.

Throughout our survey, two gaps stand out. First, no current humanoid system provides a mechanism to protect safety-critical skills from degradation caused by other updates. Second, there is no accepted benchmark for testing whether a humanoid can improve from its own deployment experience while both its environment and physical condition change.  Training a humanoid that performs well at deployment is therefore only the beginning. The harder problem is building a system that continues to improve over years of use without losing the capabilities that made it reliable in the first place. This requires more than learning from experience. A self-evolving humanoid must decide what to change, determine whether the change is actually an improvement, and remain within its safety boundaries throughout the process.

\bibliographystyle{elsarticle-harv} 
\bibliography{references}

\end{document}